\pdfoutput=1
\documentclass[runningheads]{llncs}
\usepackage[T1]{fontenc}
\usepackage{lmodern}
\usepackage{graphicx}
\usepackage{booktabs}
\usepackage{float}
\usepackage{caption}
\usepackage{multirow}
\usepackage{xcolor}
\usepackage{amsmath}
\usepackage{amsfonts}
\usepackage{amssymb}
\usepackage{tikz}
\usepackage{rotating}
\usetikzlibrary{arrows.meta,positioning,fit,backgrounds,calc}
\title{Explaining Cross-Label Decision Structure in Multi-Output Tree Ensembles}
\titlerunning{Explaining Cross-Label Decision Structure}
\author{Sylvio Barbon Junior\inst{1} \and
Hiago F. F. Oliveira\inst{2} \and
Rafael Figueira Gon\c{c}alves\inst{1} \and
Jessica Fernandes Lopes\inst{3} \and
Ricardo Cerri\inst{2}}
\authorrunning{S. Barbon Junior et al.}
\institute{University of Trieste, Trieste, Italy\\
\email{sylvio.barbonjunior@units.it} \and
Institute of Mathematics and Computer Sciences, University of S\~ao Paulo, S\~ao Carlos, Brazil\\
\email{cerri@icmc.usp.br} \and
SENAI, Londrina, Brazil}

\begin{document}
\maketitle

\begin{abstract}
A multi-output ensemble model predicts several binary labels through decision paths that it largely shares across those labels, yet the attribution methods commonly used to explain such models treat each output in isolation and therefore leave this shared structure unexamined. We propose multi-sink Decision Predicate Graphs (ms-DPGs), which extend decision predicate graphs with one sink per label, a terminal node accumulating that label's decision mass, so that each leaf contributes to the sink of every label its in-bag positive count for that label; since labels are not exclusive, these contributions need not sum to the leaf's mass. On this representation we define Shared Predicate Flow (SPF), which, for an ordered pair of labels $(A,B)$, is the share of $B$'s decision mass that passes through conditions active for both labels, and combine it with a contracted view of the graph in which label-specific lift, a label's rate in a region relative to its overall rate, indicates whether the region a shared condition delimits is one where the model assigns a higher probability to both labels, or to one at the expense of the other. Controlled experiments show that SPF captures shared decision conditions rather than mere feature overlap. On nine multi-label benchmarks, native multi-output forests share substantially more decision structure than Binary Relevance on eight, including five of the six on which the two are at predictive parity, even at equal hyperparameters, whereas per-label TreeSHAP rankings register the change of architecture without showing whether more is shared, and two case studies show how the explanation exposes which conditions the model uses for both labels of a pair. Because a native forest routes every label through the same splits, SPF describes the fitted model rather than the data, following correlated proxies and, on eight of the nine benchmarks, rising when the feature--label signal is removed; a high SPF is therefore not evidence of shared predictive information. ms-DPG thus offers a global, predicate-level account of how a multi-output ensemble model reuses decision conditions across its labels.
\keywords{Explainable AI \and Multi-label learning \and Random forests \and Decision predicate graphs}
\end{abstract}

\section{Introduction}
Multi-label classification assigns multiple labels to a single instance and can be viewed as a special case of multi-output classification in which each output is binary. We use \emph{multi-label} to refer to the prediction task and \emph{multi-output} to models that predict all labels jointly. A common baseline is Binary Relevance (BR), which fits an independent classifier for each label \cite{tsoumakas2007overview}. In contrast, a native multi-output tree selects each split using an impurity criterion aggregated across labels and stores a separate class distribution for each label at every leaf. Thus, the two approaches may achieve similar predictive performance while relying on substantially different decision structures and paths.

This difference has consequences for how their predictions can be explained. Existing work on explainable multi-label classification has largely focused on feature attribution or local rule-based explanations, including SHAP-based interpretations and methods that explain individual labels or predicted labelsets \cite{chen2021interpretation,mylonas2022lionforestsml,panigutti2020marlena}. These approaches characterize which features or rules support a prediction, but do not provide a global account of which decision predicates and paths a fitted multi-output ensemble model reuses across labels.

This limitation is particularly relevant for tree ensembles. TreeSHAP attributes a label prediction to input features \cite{lundberg2020trees}, while global feature rankings summarize importance separately for each label. Neither representation captures whether the same \emph{directional predicates}, such as $x_j>t$ rather than simply feature $x_j$, occur along decision paths supporting different labels.

Decision Predicate Graphs (DPGs) provide a representation for examining this structure directly by aggregating the decision paths of tree ensembles into graphs of predicates and their weighted connections \cite{arrighi2024dpg}. Given the feature matrix $X\in\mathbb{R}^{n\times d}$, a predicate represents a directional condition such as $x_j>t$ (we use \emph{predicate} and \emph{condition} interchangeably), where $x_j$ is the value of the $j$-th feature, $j\in\{0,\ldots,d-1\}$, and $t\in\mathbb{R}$ is a split threshold learned by the tree. Sequences of predicates form root-to-leaf paths, which we call \emph{traces}. By merging repeated predicates and their connections across trees, a DPG captures which decision conditions recur and how they are combined throughout the ensemble.

This representation can be extended naturally to the multi-output setting introduced above. Consider a tree fitted on $X$ to predict $L$ binary labels, given by $Y\in\{0,1\}^{n\times L}$. A trace containing, for example, $x_j>t_1$ and $x_k\leq t_2$ reaches a leaf $r$, which stores a positive-class probability $p_{r\ell}$ for each label $\ell\in\{0,\ldots,L-1\}$. We add to the graph one \emph{label sink} $S_\ell$ per label: a terminal node, with no outgoing connections, that collects what the leaves contribute to label $\ell$. We call this contribution \emph{decision mass}. A leaf contributes to $S_\ell$ one unit for each in-bag training sample that reaches it and is positive for $\ell$: a leaf reached by ten samples, eight of which carry label $\ell$, sends a decision mass of eight to $S_\ell$. Section~\ref{sec:sinks} shows that this mass is the forest's own output for $\ell$, accumulated over the training samples. By cross-label decision structure we mean the way a single multi-output model reuses its predicates and traces to decide several outputs at once; \emph{how the model decides} refers to this fitted structure over the training distribution, not to thresholded predictions for individual or unseen instances. Aggregating these contributions across trees exposes which predicates carry decision mass toward different label sinks, providing a global view of this structure that per-label feature importance does not capture.

Since multi-label predictions are usually explained output by output,
four research questions arise. How can a global explanation represent the
decision structure that a multi-output tree ensemble shares across its outputs
while preserving label-specific information (RQ1)? To what extent can
cross-label sharing in a fitted ensemble be quantified at the level of
directional decision conditions (RQ2)? Does predicate-level cross-label
explanation reveal structural relationships that cannot be recovered from
label association or per-label feature relevance alone (RQ3)? And how does
joint multi-output fitting affect the cross-label decision structure revealed
by the explanation, compared with independently fitted label models (RQ4)?
The paper makes four contributions, the first three tied to these questions: (i) the multi-sink Decision Predicate Graph (ms-DPG), a single predicate graph with one sink per label, which represents how far a multi-output tree ensemble decides its labels through shared predicates (RQ1); (ii) Shared Predicate Flow (SPF), the share of one label's decision mass on traces containing a predicate active for both labels of a pair, and label-specific lifts, the rate of a label among the decision mass passing through a predicate relative to its overall rate, with which a contracted ms-DPG names the shared predicates and their direction (RQ1, RQ2); (iii) formal properties of SPF, of the lifts and of the joint split criterion (Section~\ref{sec:theory}), together with an evaluation on five synthetic scenarios, in which the shared predicates are known and SPF is compared with label co-occurrence and MDI and TreeSHAP overlap (RQ3), and on nine real-world benchmarks, which compares native forests with Binary Relevance (RQ4); and (iv) a workflow for using the explanation, an assessment of it against established properties of explanation quality, and an atlas of explanations for the nine benchmarks (Section~\ref{sec:discussion} and Appendix~\ref{app:atlas}).

\section{Related Work}

\subsection{Explainability in multi-label learning}

Section~\ref{sec:mll} formalizes the multi-label approaches relevant here. Dembczy\'nski et al. distinguish marginal and conditional label dependence \cite{dembczynski2012dependence}, both of which are properties of the joint distribution of the labels. ms-DPG concerns the relations expressed through the predicates and paths of one fitted multi-output ensemble model, and high predicate sharing can arise from the model architecture or the feature representation without implying statistical dependence between the corresponding labels. Label-specific feature learning, as in LIFT \cite{zhang2015lift} and LLSF \cite{huang2015llsf}, addresses a related question at the level of the learner, by constructing or selecting features that are specific to each label or shared between correlated labels; ms-DPG instead describes the conditions that an already fitted model shares.

Existing multi-label explanation methods have largely focused on instance-level rules or feature attribution. LionForests-ML extracts local rules from multi-label random forests using per-label, labelset, and frequent-label-subset coverage strategies \cite{mylonas2022lionforestsml}, and has been adapted to multi-target regression \cite{bardos2023mtr}, while MARLENA constructs local surrogate rules for multi-label black-box classifiers \cite{panigutti2020marlena}. Tabia also addresses explainable multi-label classification \cite{tabia2019multilabel}, and SHAP-based approaches attribute individual label predictions to input features \cite{chen2021interpretation}. SEMGNN follows a different direction by incorporating explainability into a multi-label graph neural network \cite{feng2026semgnn}. These methods provide explanations for predictions or construct interpretable predictors, but they do not aggregate the directional predicates that a fitted multi-output tree ensemble reuses across label-specific outputs.

\subsection{Global, model-specific explanations of tree ensembles}

Tree ensembles admit several forms of global interpretation. Mean Decrease in Impurity (MDI) summarizes feature relevance from the impurity reductions accumulated across tree splits \cite{breiman2001random}, while TreeSHAP
provides exact feature attributions for tree models that can be aggregated across instances \cite{lundberg2017shap,lundberg2020trees}. RuleFit takes a different route by representing predictive structure through weighted
decision rules \cite{friedman2008rulefit}. For multi-target ensembles, Petkovi\'c et al. derive global feature rankings, including a structure-based symbolic score, from predictive-clustering-tree ensembles \cite{petkovic2020feature}. These methods capture feature relevance or extracted rules, but not whether different outputs reuse the same directional predicates.

Decision Predicate Graphs (DPGs) take a structural approach by aggregating repeated directional predicates and their transitions across a tree ensemble into a weighted graph \cite{arrighi2024dpg}. The resulting representation preserves relations among features, decision conditions, and model outcomes, enabling graph-based analyses such as centrality and
community detection. Subsequent work has extended this representation beyond conventional classification. Ceschin et al. adapt DPGs to Isolation Forest by tracing the contribution of predicates to inlier and outlier decisions and introducing the Inlier-Outlier Propagation Score (IOP-Score) for their global characterization \cite{ceschin2026isolation}.

The original DPG does not distinguish how predicates shared by several outputs contribute to each label; ms-DPG adds label-specific sinks and routes to each of them the decision mass of every trace for this purpose. Table~\ref{tab:related} summarizes this positioning along three
dimensions. The explanatory unit separates two groups. Feature-level methods (MDI,
TreeSHAP and the rankings of Petkovi\'c et al.) cannot tell apart two labels
that use the same feature through opposite conditions; rule- and predicate-level
methods can. Only Petkovi\'c et al.\ and ms-DPG address
several outputs, and the former still yields feature rankings, one per target
or overall. ms-DPG is therefore the only method in the table that combines a
predicate-level unit with several outputs and represents the decision
conditions shared between them, together with the direction in which they act
on each label.

\begin{table}[t]
\caption{Global explanation methods for tree ensembles: unit explained, outputs
handled, and whether shared decision conditions between outputs are represented.}
\centering
\footnotesize
\setlength{\tabcolsep}{4pt}
\begin{tabular}{@{}lllc@{}}
\toprule
Method & Explanatory unit & Outputs & Cross-output \\
\midrule
MDI \cite{breiman2001random} & feature & one at a time & -- \\
TreeSHAP \cite{lundberg2020trees} & feature & one at a time & -- \\
RuleFit \cite{friedman2008rulefit} & rule & single & -- \\
Petkovi\'c et al.\ \cite{petkovic2020feature} & feature & multi-target & -- \\
DPG \cite{arrighi2024dpg} & directional predicate & single & -- \\
IOP-Score \cite{ceschin2026isolation} & directional predicate & inlier/outlier & -- \\
ms-DPG (ours) & directional predicate & multi-label & \checkmark \\
\bottomrule
\end{tabular}
\label{tab:related}
\end{table}

\section{Multi-Label Learning with Tree Ensembles}
\label{sec:mll}

Let $\mathcal{D}=\{(\mathbf{x}_i,\mathbf{y}_i)\}_{i=1}^{n}$ with
$\mathbf{x}_i\in\mathbb{R}^d$ and $\mathbf{y}_i\in\{0,1\}^L$, where
$y_{i\ell}=1$ indicates that label $\ell\in\{0,\ldots,L-1\}$ applies to
instance $i$. A multi-label classifier estimates, for every label, a score
$\hat p_\ell(\mathbf{x})\in[0,1]$ and predicts $\hat y_\ell=1$ when it exceeds
a threshold. Label cardinality, the mean of $\sum_\ell y_{i\ell}$ over
instances, summarizes how many labels co-occur. Methods are commonly grouped
into problem-transformation approaches, which reduce the task to standard
classification problems, and algorithm-adaptation approaches, which extend a
learner to predict all labels jointly
\cite{tsoumakas2007overview,zhang2014multilabel}.

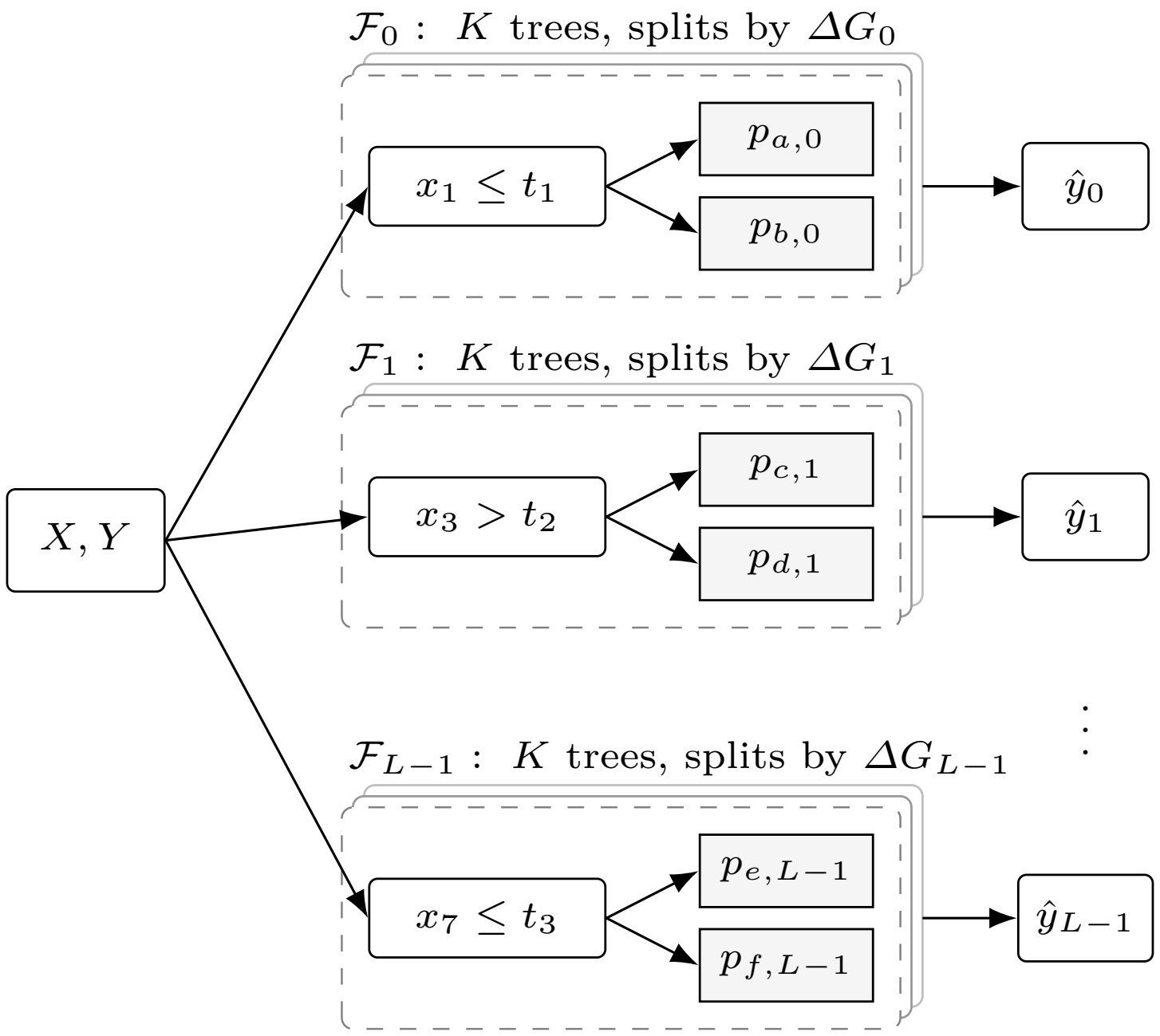
\begin{figure}[!tb]
\centering
\resizebox{.78\textwidth}{!}{%
\begin{tikzpicture}[
    >=Latex,
    font=\scriptsize,
    data/.style={draw, rounded corners=1.5pt, minimum width=10mm, minimum height=6.5mm, align=center},
    pred/.style={draw, rounded corners=1.5pt, minimum width=15mm, minimum height=5mm, align=center, fill=white},
    leaf/.style={draw, minimum width=11mm, minimum height=4.6mm, align=center, fill=black!4},
    yhat/.style={draw, rounded corners=1.5pt, minimum width=8mm, minimum height=5.5mm},
    arr/.style={->, line width=.45pt},
    group/.style={draw=black!50, dashed, rounded corners=2pt, line width=.35pt}
]
\node[data] (d) at (0,0) {$X, Y$};
\node[pred] (r0) at (2.55,2.25) {$x_1 \le t_1$};
\node[leaf] (a0) at (4.45,2.55) {$p_{a,0}$};
\node[leaf] (b0) at (4.45,1.95) {$p_{b,0}$};
\node[yhat] (o0) at (6.35,2.25) {$\hat y_0$};
\node[pred] (r1) at (2.55,.15) {$x_3 > t_2$};
\node[leaf] (a1) at (4.45,.45) {$p_{c,1}$};
\node[leaf] (b1) at (4.45,-.15) {$p_{d,1}$};
\node[yhat] (o1) at (6.35,.15) {$\hat y_1$};
\node[pred] (rl) at (2.55,-2.4) {$x_7 \le t_3$};
\node[leaf] (al) at (4.45,-2.1) {$p_{e,L-1}$};
\node[leaf] (bl) at (4.45,-2.7) {$p_{f,L-1}$};
\node[yhat] (ol) at (6.35,-2.4) {$\hat y_{L-1}$};
\node at (6.35,-1.1) {$\vdots$};
\foreach \r/\a/\b in {r0/a0/b0, r1/a1/b1, rl/al/bl}{
    \draw[arr] (d.east) -- (\r.west);
    \draw[arr] (\r.east) -- (\a.west);
    \draw[arr] (\r.east) -- (\b.west);
}
\begin{scope}[on background layer]
\foreach \r/\a/\b/\g/\name in {r0/a0/b0/g0/{\mathcal{F}_0:\ K \text{ trees, splits by } \Delta G_0},
                               r1/a1/b1/g1/{\mathcal{F}_1:\ K \text{ trees, splits by } \Delta G_1},
                               rl/al/bl/gl/{\mathcal{F}_{L-1}:\ K \text{ trees, splits by } \Delta G_{L-1}}}{
    \node[fit=(\r)(\a)(\b), inner sep=1.6mm] (\g) {};
    \draw[black!25, rounded corners=2pt, fill=white] ([xshift=4pt,yshift=4pt]\g.south west) rectangle ([xshift=4pt,yshift=4pt]\g.north east);
    \draw[black!40, rounded corners=2pt, fill=white] ([xshift=2pt,yshift=2pt]\g.south west) rectangle ([xshift=2pt,yshift=2pt]\g.north east);
    \draw[group, fill=white] (\g.south west) rectangle (\g.north east);
    \node[font=\scriptsize, anchor=south west, inner sep=1pt] at ([yshift=4pt]\g.north west) {$\name$};
}
\end{scope}
\foreach \g/\o in {g0/o0, g1/o1, gl/ol}{ \draw[arr] ([xshift=4pt]\g.east) -- (\o.west); }
\end{tikzpicture}%
}
\caption{Binary Relevance: one forest per label, each split chosen for its own
label, one class distribution per leaf. Predicates coincide across forests only
by chance. In $p_{a,0}$, the first index identifies an illustrative leaf ($a,b,c,\ldots$) and the second a label ($0,1,\ldots,L-1$), so $p_{a,0}$ is the positive-class probability of label~0 at leaf~$a$.}
\label{fig:arch-br}
\end{figure}

\paragraph{Binary Relevance.} BR, a problem-transformation approach, fits one
binary classifier per label on $(X,\mathbf{y}_{\cdot\ell})$. With random forests, each label receives its own
forest $\mathcal{F}_\ell$, and every split $s$ of a node is chosen to
maximize the impurity decrease of that label alone. We denote this decrease by
$\Delta G_\ell(s)$: the Gini impurity of label $\ell$ at the node minus the
impurities of the two children, each weighted by the fraction of the node's
samples that it receives. Each leaf stores a
single class distribution (Fig.~\ref{fig:arch-br}). Two labels can share a
decision condition only if their separately grown trees happen to select the
same split.

\begin{figure}[!tb]
\centering
\resizebox{\textwidth}{!}{%
\begin{tikzpicture}[
    >=Latex,
    font=\scriptsize,
    data/.style={draw, rounded corners=1.5pt, minimum width=10mm, minimum height=6.5mm, align=center},
    pred/.style={draw, rounded corners=1.5pt, minimum width=15mm, minimum height=5mm, align=center, fill=black!15},
    leaf/.style={draw, minimum width=29mm, minimum height=4.8mm, align=center, fill=black!4},
    yhat/.style={draw, rounded corners=1.5pt, minimum width=8mm, minimum height=5.5mm},
    arr/.style={->, line width=.45pt},
    group/.style={draw=black!50, dashed, rounded corners=2pt, line width=.35pt}
]
\node[data] (d) at (0,0) {$X, Y$};
\node[pred] (root) at (2.35,0) {$x_1 \le t_1$};
\node[pred] (mid) at (4.35,.6) {$x_3 > t_2$};
\node[leaf] (l1) at (7.05,1.0) {$(p_{a,0},\,p_{a,1},\ldots,p_{a,L-1})$};
\node[leaf] (l2) at (7.05,.2) {$(p_{b,0},\,p_{b,1},\ldots,p_{b,L-1})$};
\node[leaf] (l3) at (7.05,-.75) {$(p_{c,0},\,p_{c,1},\ldots,p_{c,L-1})$};
\node[data, align=center, minimum height=12mm] (avg) at (10.15,.12) {average over\\the $K$ trees};
\node[yhat] (o0) at (12.0,.85) {$\hat y_0$};
\node[yhat] (o1) at (12.0,.12) {$\hat y_1$};
\node[yhat] (ol) at (12.0,-.75) {$\hat y_{L-1}$};
\draw[arr] (d) -- (root);
\draw[arr] (root.east) -- (mid.west);
\draw[arr] (root.east) -- (l3.west);
\draw[arr] (mid.east) -- (l1.west);
\draw[arr] (mid.east) -- (l2.west);
\foreach \l in {l1,l2,l3}{ \draw[arr] (\l.east) -- (avg.west); }
\foreach \o in {o0,o1,ol}{ \draw[arr] (avg.east) -- (\o.west); }
\node[anchor=north west, inner sep=1pt, text=black!70] at ([yshift=-1pt]root.south west) {split: $\tfrac{1}{L}\sum_{\ell}\Delta G_\ell$};
\node[align=center, text=black!70] at (11.3,1.55) {$\hat y_\ell$ uses entry $p_{r\ell}$ of the reached leaf};
\begin{scope}[on background layer]
\node[fit=(root)(mid)(l1)(l3), inner sep=1.6mm] (gn) {};
\draw[black!25, rounded corners=2pt, fill=white] ([xshift=4pt,yshift=4pt]gn.south west) rectangle ([xshift=4pt,yshift=4pt]gn.north east);
\draw[black!40, rounded corners=2pt, fill=white] ([xshift=2pt,yshift=2pt]gn.south west) rectangle ([xshift=2pt,yshift=2pt]gn.north east);
\draw[group, fill=white] (gn.south west) rectangle (gn.north east);
\node[font=\scriptsize, anchor=south west, inner sep=1pt] at ([yshift=4pt]gn.north west) {one forest of $K$ trees shared by all labels};
\end{scope}
\end{tikzpicture}%
}
\caption{Native multi-output forest: splits chosen on the impurity decrease averaged
over labels, one class distribution per label at each leaf. All labels share
the executed predicates (shaded), which ms-DPG routes to label sinks with mass
$w_r p_{r\ell}$, where $w_r$ is the in-bag mass of leaf $r$ (Section~\ref{sec:sinks}); leaf and label indices as in Fig.~\ref{fig:arch-br}.}
\label{fig:arch-native}
\end{figure}
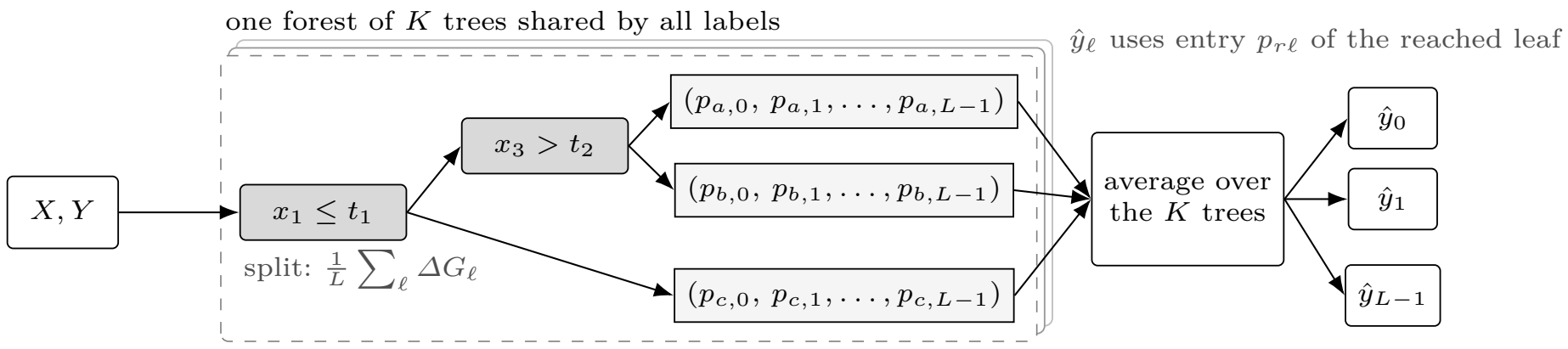

\paragraph{Native multi-output forests.} Native forests are an
algorithm-adaptation approach: multi-output decision trees, such as predictive
clustering trees \cite{blockeel1998pct} and their ensembles
\cite{kocev2013ensembles}, extend the learner to grow one tree for all labels. In the implementation
used here \cite{pedregosa2011sklearn}, a split maximizes the impurity decrease
averaged over labels, $\frac{1}{L}\sum_\ell \Delta G_\ell(s)$, and each leaf
$r$ stores one class distribution per label, from which the positive-class probability $p_{r\ell}$ of label $\ell$ at leaf $r$ is read
(Fig.~\ref{fig:arch-native}). For a forest of $K$ trees,
$\hat p_\ell(\mathbf{x})=\frac{1}{K}\sum_{k} p_{r_k(\mathbf{x}),\ell}$, where
$r_k(\cdot)$ is the function that returns the leaf of tree $k$ reached by an instance, while $r$ alone denotes a generic leaf. All labels are predicted
through the same executed predicates, and they differ only in the class
composition of the leaves.

\paragraph{Other approaches.} Label powerset, another problem-transformation approach, treats each observed labelset as one class, which models label combinations directly but multiplies the number
of classes \cite{tsoumakas2007overview}. Classifier chains append the
predictions of earlier labels to the input of later ones
\cite{read2011chains}, and multi-output tree chaining orders such a chain using
inter-target dependencies estimated from random-forest importance
\cite{mastelini2019motc}. These methods change the input or output space of
the base learner, so their predicates are not comparable across labels; BR and
native forests share learner and predicates and differ only in whether splits
are chosen per label or jointly.

\section{Multi-Sink Decision Predicate Graphs for Cross-Label Explanation}

A standard DPG aggregates the predicate structure of a tree ensemble into a weighted graph \cite{arrighi2024dpg}. In a multi-output ensemble model, however, the same trace contributes simultaneously to several label predictions. Representing these contributions requires preserving a common predicate structure while distinguishing the decision mass associated with each output. We address this requirement by extending the DPG with label-specific sinks and defining a measure of predicate sharing between outputs.

\subsection{Graph construction from executed traces}

Let $\mathcal{F}=\{T_1,\ldots,T_K\}$ be a fitted multi-output random forest with $K$ trees predicting $L$ binary labels. The ms-DPG is extracted directly from $\mathcal{F}$ without retraining or modifying the predictor. Its construction follows the execution-trace formulation of DPG, in which graph structure is derived from decision traces executed by the fitted ensemble.

For each tree $T_k$, every root-to-leaf trace reached by its in-bag training samples is recorded, weighted by the number of such samples, as stored in the fitted tree. An internal split on feature $j$ at threshold $t$ contributes the directional predicate $(j,t,\leq)$ or $(j,t,>)$ according to the branch traversed by the instance. Consecutive predicates are connected only when the corresponding transition occurs in an observed execution trace. Repeated occurrences of the same directional predicate and transition are subsequently aggregated across samples and trees, preserving their accumulated weights.

The ms-DPG also retains the executed root-to-leaf traces, so that path-level quantities are computed on paths the multi-output ensemble model actually executes and not on routes created by merging predicates from different trees.

\begin{figure}[!th]
\centering

\begin{tikzpicture}[
    >=Latex,
    node distance=9mm and 15mm,
    font=\small,
    pred/.style={
        draw,
        rounded corners=2pt,
        minimum width=24mm,
        minimum height=7mm,
        align=center,
        fill=white
    },
    leaf/.style={
        draw,
        circle,
        minimum size=8mm,
        inner sep=1pt,
        fill=white
    },
    sink/.style={
        draw,
        circle,
        minimum size=10mm,
        inner sep=1pt,
        fill=white,
        line width=1.1pt
    },
    flow/.style={
        ->,
        line width=.9pt
    },
    sinkflow/.style={
        ->,
        line width=1pt
    },
    note/.style={
        font=\scriptsize,
        align=center
    }
]

\node[pred] (p0) {$x_0 > 0.3$};

\node[pred, below left=of p0] (p1)
    {$x_2 \leq -0.1$};

\node[pred, below right=of p0] (p2)
    {$x_4 > 0.6$};

\node[pred, below=of p1] (p3)
    {$x_5 > 0.2$};

\draw[flow] (p0) -- (p1);
\draw[flow] (p0) -- (p2);
\draw[flow] (p1) -- (p3);

\node[leaf, below=8mm of p3] (r1) {$r_1$};
\node[leaf] (r2) at (p2 |- r1) {$r_2$};

\draw[flow] (p3) -- (r1);
\draw[flow] (p2) -- (r2);

\path (r1) -- (r2) coordinate[pos=.5] (mid);
\node[sink, below=25mm of mid, xshift=-14mm] (sa) {$S_A$};
\node[sink, below=25mm of mid, xshift=14mm] (sb) {$S_B$};

\draw[sinkflow] (r1) -- node[note, pos=.3, left=1pt] {$w_1p_{1A}$} (sa);
\draw[sinkflow] (r1) -- node[note, pos=.42, above=1pt, sloped] {$w_1p_{1B}$} (sb);
\draw[sinkflow] (r2) -- node[note, pos=.42, above=1pt, sloped] {$w_2p_{2A}$} (sa);
\draw[sinkflow] (r2) -- node[note, pos=.3, right=1pt] {$w_2p_{2B}$} (sb);

\node[note, above=-0.5mm of p0] (structuretitle)
    {\textbf{Executed predicate structure}};

\node[note, below=2.5mm of mid, yshift=-12mm] (sinktitle) {};
\node[note, anchor=north] at ($(sa.south)!.5!(sb.south)+(0,-1.5mm)$)
    {\textbf{Label-specific sinks}};

\begin{scope}[on background layer]
    \node[
        draw,
        dashed,
        rounded corners=3pt,
        fit=(p0)(p1)(p2)(p3)(r1)(r2)(structuretitle),
        inner sep=1.5mm
    ] (structurebox) {};

    \node[
        draw,
        dashed,
        rounded corners=3pt,
        fit=(sa)(sb),
        inner sep=1.5mm
    ] (sinkbox) {};
\end{scope}

\end{tikzpicture}

\caption{Construction of an ms-DPG. Executed traces are aggregated across trees
into shared predicates and transitions; each leaf $r_i$ sends mass
$w_i p_{i\ell}$ to the sink $S_\ell$ of label $\ell$; the contributions of one leaf to different sinks are computed label by label and need not sum to $w_i$. Here $r_1$ and $r_2$ are two example leaves, $w_i$ and $p_{i\ell}$ abbreviate $w_{r_i}$ and $p_{r_i\ell}$, and $A$ and $B$ are two labels.}

\label{fig:msdpg-concept}
\end{figure}
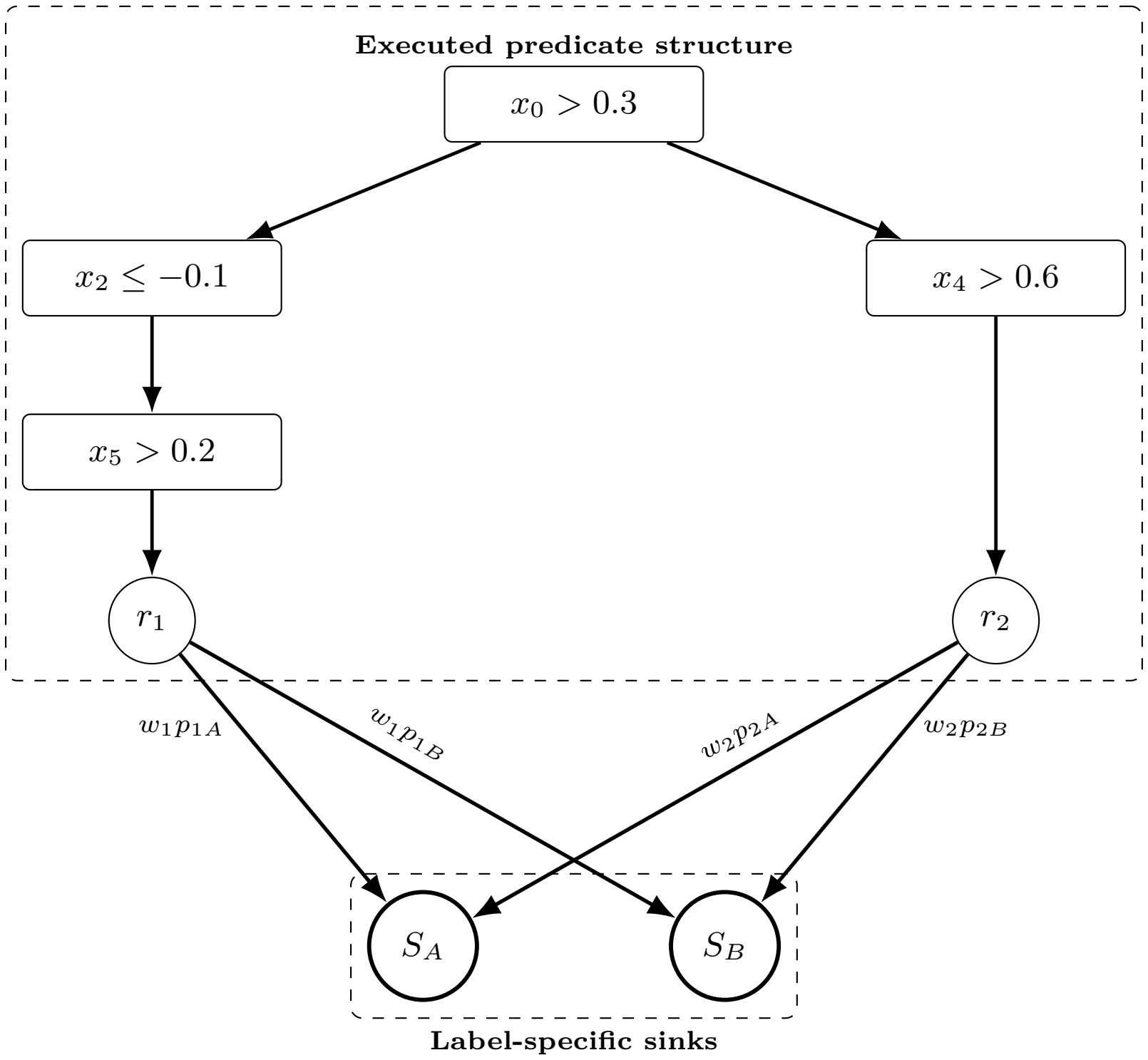

The multi-output extension occurs at the terminal leaves. A leaf contains a separate class distribution for each of the $L$ labels, so a trace may contribute differently to different outputs. Rather than constructing $L$ independent DPGs, we preserve a single predicate graph and associate each label $\ell\in\{0,\ldots,L-1\}$ with a sink $S_\ell$, a terminal node with no outgoing connections in which all the mass directed to that label accumulates. Each trace is then connected to these sinks through its label-specific fractional contributions, defined below. The resulting ms-DPG combines a shared, execution-respecting predicate structure with output-specific destinations, as illustrated in Fig.~\ref{fig:msdpg-concept}.

\subsection{Label sinks and decision mass}
\label{sec:sinks}

As in Section~\ref{sec:mll}, $r$ denotes a leaf and $p_{r\ell}$ the positive-class fraction stored at $r$ for label $\ell$. Let $w_r$ be the leaf mass of $r$, defined as the sum of the weights of the training instances reaching $r$. When all instances have unit weight, $w_r$ reduces to the number of in-bag instances in the leaf, counted with their bootstrap multiplicity. The contribution of leaf $r$ to sink $S_\ell$ is defined as
\begin{equation}
    m(r,S_\ell)=w_rp_{r\ell}.
    \label{eq:fractional-mass}
\end{equation}

The same leaf can thus contribute different amounts of mass to different label sinks. For two labels $A$ and $B$, for example, a leaf with mass $w_r$ contributes $w_rp_{rA}$ to $S_A$ and $w_rp_{rB}$ to $S_B$. These contributions retain the label-specific class composition of the leaf while preserving the shared trace that produced it.

\paragraph{Relation to the ensemble prediction.} Because $p_{r\ell}$ is the in-bag positive fraction at $r$, $w_rp_{r\ell}$ is both the in-bag positive count of $\ell$ at $r$ and $\sum_{i:\,r_k(\mathbf{x}_i)=r}c_{ki}\,p_{r_k(\mathbf{x}_i),\ell}$, where the sum runs over the training instances routed to leaf $r$ of tree $T_k$ (the condition $r_k(\mathbf{x}_i)=r$ uses the leaf-reaching function of Section~\ref{sec:mll} and does not redefine $r$), and $c_{ki}$ is the bootstrap count of instance $i$ in $T_k$. Sink $S_\ell$ therefore accumulates the summands of $\hat p_\ell$ (Section~\ref{sec:mll}), up to the factor $1/K$, over the in-bag sample of each tree; the factor cancels in Shared Predicate Flow (SPF, Section~\ref{sec:spf}) and in the lifts (Section~\ref{sec:read}), which are ratios of masses. The structure learned by the trees, the decision mass of $\ell$ along its traces and the tree-level probabilities from which the ensemble forms its prediction thus coincide by construction. ms-DPG is a global account of the fitted model weighted by the training distribution: it does not decompose individual predictions, does not depend on the threshold applied to $\hat p_\ell$, and does not describe the model's behaviour on unseen data.

\subsection{Shared Predicate Flow: quantifying cross-label sharing}
\label{sec:spf}

The multi-sink representation identifies where label-specific decision mass flows through the graph. We use this structure to quantify the extent to which the predicates associated with one label also occur on traces contributing to another. Shared Predicate Flow (SPF) measures this relationship for an ordered pair of labels.

For each label $\ell$, a split $s$ at an internal node $u$ of in-bag mass $w_u$ produces two directional predicates, one per child. Its label-specific gain $\Delta G_\ell(s)$ (Section~\ref{sec:mll}) is computed from the class fractions stored for $\ell$, even though the split was selected on the criterion averaged over labels, and is credited to the child in which the positive-class fraction of $\ell$ is higher, half to each child when the fractions coincide. With $\delta_\ell(u,v)\in\{0,\frac12,1\}$ the share credited to predicate $v$, the aggregated gain of $v$ for $\ell$ is
\begin{equation}
g_\ell(v)=\sum_{u}\delta_\ell(u,v)\,w_u\,\Delta G_\ell(s_u),
\label{eq:gain}
\end{equation}
where the sum runs over the internal nodes $u$ of all trees whose split $s_u$ produces a predicate equivalent to $v$. A predicate is \emph{active} for $\ell$, written $v\in\mathcal{A}_\ell$, when $g_\ell(v)$ lies within the top $25\%$ of the positive values of $g_\ell$ and is at least $5\%$ of their maximum. The first criterion retains the predicates that are prominent for the label, the second excludes those whose contribution is negligible. These cut-offs are heuristic rather than derived from a statistical criterion; within the ranges tested in Section~\ref{sec:res-robust}, their choice does not change the conclusions.

For an ordered pair of labels $(A,B)$, let $C_{A,B}=\mathcal{A}_A\cap\mathcal{A}_B$ denote the
predicates active for both labels and $\mathcal{P}_B$ the executed
root-to-leaf traces contributing positive mass to sink $S_B$.
For each trace $p\in\mathcal{P}_B$, we define the indicator function
\begin{equation}
\mathbb{1}_{A,B}(p)=
\begin{cases}
1, & \text{if $p$ contains a predicate in $C_{A,B}$},\\
0, & \text{otherwise.}
\end{cases}
\end{equation}

If $m_p$ denotes the $B$-specific mass carried by trace $p$, Shared
Predicate Flow is
\begin{equation}
\mathrm{SPF}(A\to B)=
\frac{\sum_{p\in\mathcal{P}_B} m_p\,\mathbb{1}_{A,B}(p)}
     {\sum_{p\in\mathcal{P}_B} m_p}.
\label{eq:spf}
\end{equation}

SPF is the fraction of the decision mass toward $B$ carried by executed traces that contain at least one active predicate shared with $A$. It lies in $[0,1]$, where $0$ means that no target mass traverses a shared active predicate and $1$ that all of it does. SPF is set to 0 when $B$ has no positive mass and when $C_{A,B}$ is empty, which includes the case of a label without active predicates; pairs with $A=B$ are not considered, so a forest with $L$ labels has $L(L-1)$ ordered pairs. The arrow fixes which label's mass forms the denominator; it does not indicate an influence of $A$ on $B$.

\paragraph{Coverage rather than degree.} SPF is a coverage measure: a trace with a single shared active predicate counts in full, as does one whose active predicates are all shared, so SPF quantifies how much of $B$'s mass lies on traces with at least one predicate active for both labels, not how much of each trace is shared. We use the indicator because a trace is a conjunction: the probability its leaf assigns to $B$ is conditional on every predicate on it, so one shared predicate already makes that part of $B$'s output depend on a condition the model also uses for $A$. A measure of the degree of sharing, such as the fraction of each trace's predicates that are shared, would also depend on tree depth, since most predicates on a trace are active for no label, and a comparison between architectures of different depth would partly compare depths. A depth-free variant leads to the same conclusions (Appendix~\ref{app:analyses}). How much of a trace is shared is shown instead by the contracted ms-DPG (Section~\ref{sec:read}), which names the shared predicates and the transitions between them.

\paragraph{SPF for Binary Relevance.} A BR model consists of $L$ independent
forests $\mathcal{F}_0,\ldots,\mathcal{F}_{L-1}$. We merge them into a single
ms-DPG in which every trace of $\mathcal{F}_\ell$ is connected only to sink
$S_\ell$, with mass $w_r p_r$ given by the leaf weight and the positive-class
fraction of its single label. Gains, and hence the predicates active for $\ell$,
are computed from $\mathcal{F}_\ell$ alone. Predicates of different forests are
identified by the same key as within a native forest, namely feature,
direction and threshold bin, so that $C_{A,B}$ contains the predicates that the
forests of $A$ and $B$ select independently. Eq.~\ref{eq:spf} then applies
unchanged, with $\mathcal{P}_B$ the traces of $\mathcal{F}_B$: SPF$(A\to B)$ is
the fraction of $B$'s decision mass on traces of $\mathcal{F}_B$ that contain a
predicate also active in $\mathcal{F}_A$.

\subsection{Reading the explanation: contracted views and label-specific lifts}
\label{sec:read}

The method produces two complementary views. The matrix of SPF values ranks ordered label pairs by the fraction of the target label's decision mass that travels through predicates active for the source label, and so indicates \emph{where} the fitted forest reuses decision conditions across outputs. For a selected pair, a contracted ms-DPG then shows the conditions themselves. It keeps the co-enriching shared active predicates (defined below) with the largest product of mass and smaller lift, so that broad conditions traversed by most samples are not preferred, orders them by their depth in the trees, links them by the executed transitions between them, groups the leaves into four classes by the labelset they store for the pair, and connects these classes to the label sinks. With fully grown trees the leaves are pure, so a leaf's class is the labelset of the in-bag instances that reach it. Class nodes thus describe tree-level outputs, not ensemble predictions: an instance's prediction averages the leaves it reaches in all $K$ trees, which need not share a class.

Since mass alone does not indicate whether a shared predicate favours a label, each predicate $v$ is annotated with its label-specific lift. Let $W(v)$ be the leaf mass $\sum w_r$ of the traces that contain $v$, each trace counted once, $F_\ell(v)$ the decision mass for $\ell$ on these traces, and $\pi_\ell=\sum_r m(r,S_\ell)/\sum_r w_r$ the forest-wide rate of $\ell$, with both sums over all leaves of the forest. Then
\begin{equation}
\lambda_\ell(v)=\frac{F_\ell(v)/W(v)}{\pi_\ell}.
\label{eq:lift}
\end{equation}
The lift compares the rate of $\ell$ among the mass passing through $v$ with its rate in the whole forest or, equivalently by the relation in Section~\ref{sec:sinks}, the mean tree-level probability for $\ell$ on these traces with its overall mean. A lift above 1 means that $\ell$ is enriched in this region, and a lift below 1 that it is depleted; neither is an incremental or causal effect of the split. A shared predicate is \emph{co-enriching} when $\lambda_A,\lambda_B>1$, i.e.\ the traces through it are enriched in both labels relative to the whole forest, and \emph{separating} when one lift exceeds 1 and the other falls below it.

\paragraph{Global enrichment, not local effect.} The lift describes the region of the input space delimited by the traces through $v$, including the conditions tested above it; it is not the incremental effect of the split that produces $v$ relative to its parent node. Its local counterpart, the change in the rate of the label from the parent node to the children reached through $v$, mostly exceeds 1 for active predicates, since a predicate's gain is credited to the child in which the label's rate increases; Appendix~\ref{app:analyses} compares the two readings. ``Co-enriching'' and ``separating'' therefore describe where a condition places the instances relative to the whole training distribution, not whether the split itself moves the prediction up or down. Table~\ref{tab:reading} lists, for each component, the question it answers about a multi-output forest. Section~\ref{sec:res-emotions} provides a real-data example.

\begin{table}[t]
\caption{Reading guide for an ms-DPG explanation.}
\centering
\scriptsize
\setlength{\tabcolsep}{3pt}
\renewcommand{\arraystretch}{1.15}
\begin{tabular}{@{}p{2.2cm}p{9.6cm}@{}}
\toprule
Component & Question it answers \\
\midrule
SPF matrix & Which label pairs does the forest decide through common conditions? \\
Predicate node & Which conditions carry decision mass toward both labels? \\
Lift $\lambda_\ell$ & Is the predicted rate of $\ell$ on the traces through the condition above ($\lambda_\ell>1$) or below ($\lambda_\ell<1$) its forest-wide rate? \\
Lifts of a pair & Are these traces enriched in both labels (co-enriching) or in one and depleted in the other (separating)? \\
Transition arc & In which order do the trees test the conditions? \\
Class node & Which labelset for the pair do the leaves reached through these conditions store? \\
Label sink & How much of a label's decision mass is explained? \\
Omitted node & How much of each class does the view leave out? \\
\bottomrule
\end{tabular}
\label{tab:reading}
\end{table}

\section{Formal Properties of the Explanation}
\label{sec:theory}

This section supports the answers to RQ1--RQ4 with exact results. Propositions~\ref{prop:mass}--\ref{prop:lift} hold for any fitted forest or for the explicit constructions given; Proposition~\ref{prop:selection}, and Proposition~\ref{prop:d1} in Appendix~\ref{app:theory}, concern split selection under stated population assumptions. They describe the measures and the fitting criterion, and are not a general claim about the interpretability or faithfulness of the explanation. Full proofs are in Appendix~\ref{app:proofs}.

We assume unit instance weights and keep the notation of Section~\ref{sec:sinks}. With $\pi_\ell$ the forest-wide rate of Eq.~\ref{eq:lift}, let $\bar v$ be the mirror of a predicate $v$: same feature and same threshold or threshold bin, opposite direction.

\begin{lemma}[Binary Gini gain]
\label{lem:gini}
If a node is split into two non-empty children that receive fractions $P_L,P_R$ of its mass and have positive rates $p_L,p_R$ for label $\ell$, then $\Delta G_\ell=2P_LP_R(p_L-p_R)^2$. Hence $\Delta G_\ell\geq0$, with equality if and only if $p_L=p_R$, and $\Delta G_\ell$ is unchanged when $y_\ell$ is replaced by $1-y_\ell$.
\end{lemma}

\subsection{Label-wise mass accounting (RQ1)}

\begin{proposition}[Label-wise mass conservation and prediction recovery]
\label{prop:mass}
For a native forest and every label $\ell$: (i) $\sum_r m(r,S_\ell)=\sum_k\sum_i c_{ki}\,y_{i\ell}$, the total in-bag positive count; (ii) for every input $\mathbf{x}$,
\[\hat p_\ell(\mathbf{x})=\frac{1}{K}\sum_k m(r_k(\mathbf{x}),S_\ell)/w_{r_k(\mathbf{x})};\] (iii) if the $\ell$-flow on an arc is the sum of $w_rp_{r\ell}$ over all traversals of that arc, counted with multiplicity, where each trace enters from its tree root and exits to $S_\ell$, then inflow equals outflow at every predicate node.
\end{proposition}

With the retained trace index, label-specific information is therefore not lost when predicates are merged: each sink accounts for all of the decision mass of its label and, given the leaves an input reaches, the leaf masses reproduce the forest's probabilities exactly. The sink layer is the only addition to the predicate structure, so for $L=1$ predicates and transitions are those of the DPG, as verified against the reference implementation in Section~\ref{sec:experiments}.

\subsection{Range and bounds of SPF (RQ2)}

\begin{proposition}[Range, extremes and bounds]
\label{prop:range}
For every ordered pair $(A,B)$ with $\sum_{p\in\mathcal{P}_B}m_p>0$: (i) $0\leq\mathrm{SPF}(A\to B)\leq1$; (ii) $\mathrm{SPF}(A\to B)=0$ if and only if no trace in $\mathcal{P}_B$ contains a predicate of $C_{A,B}$, which holds in particular when $C_{A,B}=\emptyset$; (iii) $\mathrm{SPF}(A\to B)=1$ if and only if every trace in $\mathcal{P}_B$ contains a predicate of $C_{A,B}$; (iv) if $\ell_p$ is the length of trace $p$ and $h_p$ the number of its predicate occurrences in $C_{A,B}$, with $h_p/\ell_p:=0$ when $\ell_p=0$, then
\[
\frac{\sum_{p} m_p\,h_p/\ell_p}{\sum_{p} m_p}\;\leq\;\mathrm{SPF}(A\to B)\;\leq\;\min\Bigl\{1,\;\frac{\sum_{p} m_ph_p}{\sum_{p} m_p}\Bigr\}.
\]
\end{proposition}

Appendix~\ref{app:theory} uses item (iv) to relate SPF to the density of shared predicates on a trace and to the length of the traces, which helps to read the permutation results of Section~\ref{sec:res-arch}.

\subsection{Limits of feature-level attributions (RQ3)}

\begin{proposition}[Label complement]
\label{prop:complement}
Replace $y_B$ by $1-y_B$ in the training data. (i) The split criterion of every candidate split is unchanged, for native and BR forests alike, so the same random stream yields the same trees, given deterministic tie-breaking as in scikit-learn. (ii) Every gain of $B$ moves to the mirror predicate: the aggregated gains satisfy $g'_B(\bar v)=g_B(v)$, so $v$ is active for $B$ after the change if and only if $\bar v$ was active before. (iii) The per-label MDI profile of $B$, its mean $|$TreeSHAP$|$ profile and $|\phi(y_A,y_B)|$ are unchanged for every $A\neq B$.
\end{proposition}

Summaries that ignore direction are therefore blind to this change, whereas $C_{A,B}$ can change completely.

\begin{proposition}[SPF is not determined by feature-level summaries or by $Y$]
\label{prop:dissociation}
(a) There exist two label configurations, on the same inputs and the same forest structure, for which the per-label MDI profiles, the per-label mean $|$TreeSHAP$|$ profiles and $|\phi|$ all coincide, whereas $\mathrm{SPF}(A\to B)$ equals 1 in one and 0 in the other. (b) There exist a label matrix $Y$ and two feature matrices such that forests of stumps without bootstrap, each selecting a split that maximizes the averaged criterion, have $\mathrm{SPF}(A\to B)=1$ on one and $0$ on the other.
\end{proposition}

Part (a) is a minimal form of scenario D5 (Section~\ref{sec:synth}), and part (b) shows that label association alone cannot fix SPF.

\subsection{Properties of the lifts}

\begin{proposition}[Lift identities]
\label{prop:lift}
Assume $\pi_A,\pi_B,\pi_\ell>0$, so that the lifts are defined. Let $F_{AB}(v)$ and $F_{\emptyset}(v)$ be the in-bag mass, on traces containing $v$, of instances positive for both $A$ and $B$ and for neither. (i) $\pi_A\lambda_A(v)+\pi_B\lambda_B(v)=1+\bigl(F_{AB}(v)-F_{\emptyset}(v)\bigr)/W(v)$. (ii) If $A$ and $B$ never co-occur in the training data, then $\pi_A\lambda_A(v)+\pi_B\lambda_B(v)\leq1$, so a co-enriching predicate requires $\pi_A+\pi_B<1$ and $\min\{\lambda_A(v),\lambda_B(v)\}\leq1/(\pi_A+\pi_B)$. (iii) If $V$ is a set of predicates such that every trace contains exactly one element of $V$, then $\sum_{v\in V}W(v)\lambda_\ell(v)=\sum_r w_r$.
\end{proposition}

By (iii), enrichment in some regions is balanced by depletion in others, and by (ii) co-enrichment is constrained when labels are mutually exclusive.

\subsection{Joint split selection and shared predicates (RQ4)}

The native criterion ranks a split by the sum of its gains over labels, so a split can be selected although it is not the best for any single label.

\begin{proposition}[Joint criterion favours shared conditions]
\label{prop:selection}
Let $\sigma_s$, $\sigma_{A,1},\ldots,\sigma_{A,m}$ and $\sigma_{B,1},\ldots,\sigma_{B,m}$, $m\geq1$, be independent uniform $\pm1$ features and, for $\ell\in\{A,B\}$, let $P(y_\ell=1\mid\sigma)=\frac12+\alpha\sigma_s+\beta\sum_i\sigma_{\ell,i}$, with $\alpha>0$, $|\beta|>\alpha>|\beta|/\sqrt2$ and $\alpha+m|\beta|\leq\frac12$. Let trees evaluate the criterion on the population, with all features as candidates at every split. Then: (i) the root of every native tree splits on $\sigma_s$; (ii) in BR, every tree of label $\ell$ with depth at most $m$ (a stump has depth 1) splits only on $\sigma_{\ell,1},\ldots,\sigma_{\ell,m}$, so $C_{A,B}=\emptyset$ and $\mathrm{SPF}(A\to B)=0$; (iii) at depth 1, native forests have $\mathrm{SPF}(A\to B)=\frac12+\alpha$.
\end{proposition}

The proposition explains how the joint criterion can bring a shared condition forward. It holds on the population and at limited depth, and does not account for the size of the excess observed in fully grown trees on the real benchmarks (Section~\ref{sec:res-real}). Appendix~\ref{app:theory} states the corresponding result for scenario D1, discusses larger depths and reports numerical checks of all propositions.

\section{Experimental Setup}
\label{sec:experiments}

We compare native multi-output random forests with Binary Relevance (BR)
on synthetic diagnostic scenarios and real-world multi-label benchmarks.
Both approaches use 200-tree \texttt{RandomForestClassifier} models, with
BR hyperparameters selected from a fixed validation grid to match the
validation macro-F1 of the native forest. The grid crosses a maximum depth of 8, 12, 16 or unlimited, a minimum of 1, 2 or 5 instances per leaf, and $\sqrt{d}$ or $d/2$ candidate features per split (24 configurations); the native forest keeps the defaults (unlimited depth, one instance per leaf, $\sqrt{d}$ features), and the configuration with the smallest validation gap is kept. A BR model thus consists of 200 trees per label, $200L$ in total, against 200 shared trees in the native forest. Labels are predicted positive when their estimated probability exceeds .5, the default decision rule of the implementation. Results are reported over ten
random repetitions, each a different random 60/20/20\% partition into training, validation and test sets, used by both architectures. In the result plots, blue denotes the native forest, orange BR,
and grey baselines, nulls and thresholds. Native and BR forests are compared
at predictive parity, defined as a median absolute test macro-F1 gap over the ten repetitions of at most .015; a comparison that meets this tolerance at the median need not meet it in every repetition, and the number of repetitions that do is reported with the results.

The experiments address the research questions as follows. Section~\ref{sec:res-matching} examines how the definition of predicate equivalence affects structural comparisons across forests; it is a methodological check of the matching rule on which the other analyses rely, not a research question of its own. Section~\ref{sec:res-synth} asks whether SPF responds to shared decision conditions rather than to shared features or label association (RQ3). Sections~\ref{sec:res-real} and~\ref{sec:res-arch} compare native forests with BR at comparable predictive performance and under label-permutation nulls (RQ4). Section~\ref{sec:res-robust} examines how these comparisons depend on the matching rule and the activity cut-offs and, with the definition of SPF (Section~\ref{sec:spf}), answers RQ2. Section~\ref{sec:res-shap} contrasts this with the view of per-label feature attributions (RQ3), and Sections~\ref{sec:res-emotions} and~\ref{sec:res-flags} apply the contracted ms-DPG to label pairs of two benchmarks (RQ1).

SPF is computed for every ordered label pair. Predicates are matched across
trees by feature, direction and threshold bin: each threshold is mapped to one
of $b=32$ feature-wise quantile bins of the training data, so that nearby
thresholds represent the same
decision region. Exact matching, which requires thresholds equal to six decimal places, serves
as a contrast in Section~\ref{sec:res-matching}.

The ms-DPG is computed directly from the arrays of the fitted trees. On a
single-output forest of 200 trees it reproduces the predicate nodes and
transition counts of the reference DPG implementation\footnote{\url{https://github.com/Meta-Group/DPG}}
exactly, and SPF is invariant under label duplication. For a forest with $N$
nodes, $R$ leaves and root-to-leaf traces of total length $\Lambda$, extraction takes
$O(LN\log b+L\Lambda)$ time and $O(LR+\Lambda+LQ)$ memory for $Q$ distinct predicates, and
SPF for all ordered pairs is computed in a single pass over the traces in
$O(L\Lambda+L^2R)$ time. On \emph{yeast} with all 14 labels (200 trees, about
130{,}000 leaves, one CPU core), extraction took 16.8\,s and SPF for the 182
ordered pairs 12.9\,s, against 4.1\,s for training the forest. The implementation, the protocols of the analyses and the scripts that produce every table and figure are available at \url{https://github.com/sbarbonjr/ms-dpg}.

\subsection{Synthetic scenarios with known decision structure}
\label{sec:synth}

The synthetic scenarios were designed to isolate specific sources of apparent cross-label sharing and to test whether SPF responds to genuinely shared decision structure rather than to correlation, independence, hierarchical dependence, or common feature use alone. Each synthetic scenario contains 1,500 instances, 50 features, and four binary labels. Unless otherwise specified, features are independently drawn from $\mathcal{N}(0,1)$. Labels 0 and 1, whose binary values are written $y_0$ and $y_1$, form the diagnostic pair, while labels 2 and 3 serve as independent controls, with $y_2=1$ for $x_{10}>0$
and $y_3=1$ for $x_{11}>0.3$, and $0$ otherwise. Each label is independently flipped with probability $0.05$. For each of the ten repetitions, the data are divided into 60\% training, 20\% validation, and 20\% test sets.

\begin{figure}[t]
\centering
\includegraphics[width=\textwidth]{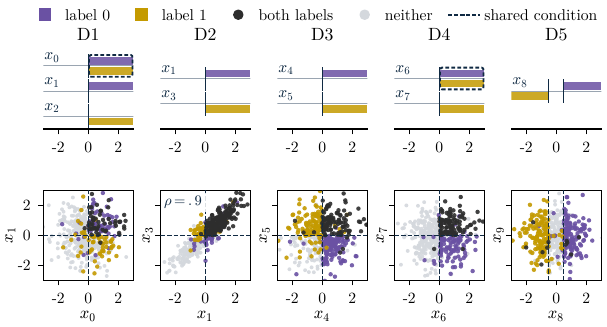}
\caption{Synthetic scenarios. Top: conditions defining label~0 (purple) and
label~1 (ochre); dashed outlines mark a predicate shared by both. Bottom: 450
training instances of one repetition by label combination, with the planted
thresholds dashed. $x_9$ in D5 is irrelevant. Label~0 and label~1 are the binary outputs $y_0,y_1\in\{0,1\}$ of the text, and the axes $x_j$ are input features.}
\label{fig:scenarios}
\end{figure}

The five scenarios, depicted in Fig.~\ref{fig:scenarios}, examine different
conditions under which cross-label predicate sharing may arise. D1 provides a positive reference case:
$y_0=1$ when $x_0>0$ and $x_1>0$, whereas $y_1=1$ when $x_0>0$ and
$x_2>0$, so that the two labels share the planted directional predicate
$x_0>0$.

D2 examines the effect of correlated predictors without a shared generating
predicate. Here, $y_0=1$ when $x_1>0$ and $y_1=1$ when $x_3>0$, with
$x_3=0.9x_1+\sqrt{0.19}\epsilon$ and
$\epsilon\sim\mathcal{N}(0,1)$ independent of $x_1$. This construction
gives $\operatorname{Var}(x_3)=1$ and
$\operatorname{corr}(x_1,x_3)=0.9$. The scenario tests whether the fitted
forest develops shared predicate structure through correlated proxies when the
planted rules use different features.

D3 provides an independent null case: $y_0=1$ when $x_4>0$ and $y_1=1$
when $x_5>0$. Since the two labels depend on distinct and independent
generating features, cross-label predicate sharing is not expected.

D4 introduces hierarchical structure. Label 0 is positive when $x_6>0$,
while label 1 is positive only when $y_0=1$ and $x_7>0$. The constraint
$y_1\leq y_0$ is enforced after label noise, so the child label necessarily
inherits the parent condition, so that shared structure is expected.

D5 distinguishes predicate sharing from feature overlap. Both labels depend
on $x_8$, but in disjoint regions: $y_0=1$ when $x_8>0.5$ and $y_1=1$
when $x_8<-0.5$. Label-specific split gain is assigned to the branch with
the higher positive-class rate, preserving the two regions as distinct
directional predicates. The scenario tests whether SPF responds to shared
decision conditions or merely to common feature use.

\subsection{Multi-label benchmarks}
\label{sec:data}

The real-world evaluation uses nine multi-label benchmarks from OpenML~\cite{vanschoren2014openml}, each pinned by name and version: \emph{emotions}, based on timbral and rhythmic audio features of music
excerpts~\cite{trohidis2008emotions}; \emph{scene}, representing
natural images; \emph{yeast}, describing gene-function annotations; \emph{image}, natural scenes described by colour features~\cite{zhang2007mlknn}; \emph{birds}, bird species present in audio recordings~\cite{briggs2013birds}; \emph{genbase}, protein families described by motifs~\cite{diplaris2005protein}; \emph{enron}, categories of e-mail messages~\cite{klimt2004enron}; and \emph{slashdot} and \emph{langlog}, subject categories of technology news items and of posts on a linguistics blog~\cite{read2011chains}. The last three are text data with about 1,000 binary word features. Their characteristics are reported in Table~\ref{tab:datasets}. Numeric features are kept, with missing values imputed by the column median; categorical features with two values are encoded as 0/1 and other categorical features are dropped, as are features constant over the dataset (1,073 of the 1,185 of \emph{genbase}). None of the nine benchmarks has missing values, so the imputation has no effect. A feature that is constant over the whole dataset is constant in every training set and can never be split on, so removing it uses no information about the test data. The quantile bins used for predicate matching are computed on the training set of each repetition. For \emph{yeast}, and for every benchmark with more than eight labels, we retain the eight most prevalent labels, and instances without any retained label are kept; Section~\ref{sec:res-real} reports all 14 labels of \emph{yeast} and, for the other benchmarks with more than eight labels, every label with at least 20 positive instances, as sensitivity analyses. The benchmarks differ in domain, dimensionality, number of labels, and label cardinality, providing nine settings in which the cross-label decision structure is not known a priori. They range from almost single-label data (\emph{scene}, \emph{genbase}) to dense labelsets (\emph{yeast}); we do not claim that this selection represents multi-label data in general (Section~\ref{sec:limitations}).

A fourth dataset, \emph{flags} from the UCI repository (194 national flags),
illustrates the explanation on features that need no domain knowledge and is
not part of the quantitative comparison. Its multi-label version takes the
seven colours present on a flag as labels and the descriptive attributes as
features, with landmass, zone, language and religion one-hot encoded; the
number of colours and the colour-derived attributes (main hue, top-left and
bottom-right colour) are excluded, since they reveal the labels.

\begin{table}[!th]
\caption{Real-world benchmarks. $n$: instances; $d$: features after preprocessing; $L$: labels retained (all labels in parentheses);
Card.: mean number of positive labels per instance, over the retained labels.}
\centering
\footnotesize
\begin{tabular}{llrrrr}
\toprule
Dataset & Domain & $n$ & $d$ & $L$ & Card.\\
\midrule
emotions & Music & 593 & 72 & 6 & 1.87\\
scene    & Images & 2407 & 294 & 6 & 1.07\\
yeast    & Gene function & 2417 & 103 & 8 (14) & 3.55\\
image    & Images & 2000 & 135 & 5 & 1.24\\
birds    & Audio & 645 & 259 & 8 (19) & 0.73\\
genbase  & Proteins & 662 & 112 & 8 (27) & 0.90\\
enron    & E-mail & 1702 & 1001 & 8 (53) & 2.30\\
slashdot & Text & 3782 & 1079 & 8 (22) & 0.87\\
langlog  & Text & 1460 & 998 & 8 (75) & 0.53\\
\bottomrule
\end{tabular}
\label{tab:datasets}
\end{table}

\subsection{Baseline explanations and controls}
\label{sec:baselines}

To distinguish shared predicate structure from simpler forms of cross-label
similarity, SPF is compared with three diagnostic baselines. Label
co-occurrence measures association directly from the observed binary
outputs, as the absolute phi coefficient $|\phi|$ between the two label columns. MDI overlap compares impurity-based feature-importance profiles
derived from the fitted forests \cite{breiman2001random}, while SHAP overlap
compares feature-attribution profiles obtained with TreeSHAP
\cite{lundberg2020trees}. They test whether high SPF can be explained by co-occurrence or shared
feature importance alone.

Two controls complete the design: label-permutation nulls, which fit the
native forest to training labels permuted independently per column,
jointly by row, or per column within deciles of an out-of-fold score, so as to
remove both label dependence and feature--label signal, only the signal, or
only the dependence, and an
equal-hyperparameter control, which fits it with the hyperparameters
selected for BR. Per-label TreeSHAP rankings of two forests are compared by
Spearman correlation on 200 test instances (all test instances on \emph{birds} and \emph{genbase}, which have fewer).

Path-dependent TreeSHAP is numerically unstable on very deep trees. The native trees of \emph{enron} and \emph{slashdot} have mean depths of about 60 and 240 levels, and on them the SHAP values of some labels do not add up to the model output, with errors of up to $1.4\times10^{4}$ on \emph{enron} and $10^{62}$ on \emph{slashdot}, although the tree view passed to TreeSHAP reproduces the forest output to $10^{-15}$. We therefore verify additivity for every forest and label with the tolerance of the reference TreeSHAP implementation, compare rankings only over the labels that satisfy it in both forests, and do not compute forests deeper than 100 levels. This rule was adopted after the instability had been observed and is therefore a post hoc choice. On \emph{emotions}, \emph{scene} and \emph{yeast}, whose native trees are at most 27 levels deep, every label of the native and BR forests of the first repetition satisfies the criterion.

\section{Results}
\label{sec:results}

\subsection{Effect of predicate matching}
\label{sec:res-matching}

Predicate equivalence determines the outcome of structural comparisons. In
D1, where labels 0 and 1 share the planted condition $x_0>0$, exact threshold
matching yields median differences in SPF between the native forest and Binary Relevance (native $-$ BR) of .233 for $0\to1$ and
.236 for $1\to0$. With matching by 32 quantile bins the differences almost vanish (.016 and $-.022$), all bootstrap
intervals include zero, and median
SPF is approximately .49--.50 for the native forest and .50--.52 for BR. Both architectures recover the planted rule when
nearby thresholds are treated as equivalent, whereas exact matching
exaggerates the difference between them.

\subsection{Shared predicates versus shared features and label association (RQ3)}
\label{sec:res-synth}

\begin{figure}[!htbp]
\centering
\includegraphics[width=\textwidth]{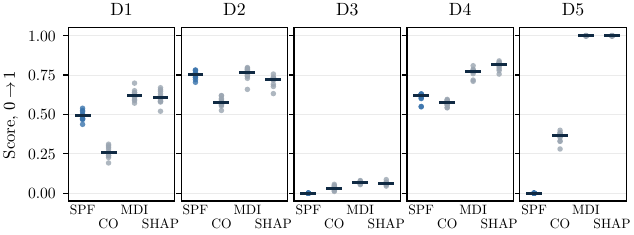}
\caption{Scores for the pair $0\to1$ in the synthetic scenarios: native SPF
(blue) and baselines (grey): CO, label co-occurrence $|\phi|$; MDI and SHAP,
cosine similarity of the two labels' importance profiles. Points are
repetitions, segments medians.}
\label{fig:synthetic}
\end{figure}

Across the synthetic scenarios, SPF responds to shared decision conditions and
not to label association or feature overlap as such
(Fig.~\ref{fig:synthetic}). The contrast is clearest in D5, where both labels
depend on $x_8$ but in opposite regions of its domain. MDI and SHAP report
almost complete feature overlap, while SPF is zero because the two labels use
distinct directional predicates; a shared predictive feature is not a shared
decision condition.

D2 illustrates the converse situation. Although its planted rules use
different features, $x_1$ and $x_3$ are strongly correlated, and the native
forest reaches an SPF of .751 for $0\to1$ and .758 for $1\to0$. Label~1 still
assigns a larger median gain to its generating feature $x_3$ than to the proxy
$x_1$ (38,074 versus 18,061 in the native forest; 38,406 versus 17,333 under
BR), so the high SPF reflects the predicates used by the fitted forest,
proxies included, and not the data-generating rule.

In D3 all measures are close to zero. In D4 the child label inherits the
parent condition and SPF is high in both directions (.62 and .60), so that it
records the shared condition but not which label contains the other.
The fraction of the child's decision mass on traces through predicates that support the parent, a containment measure defined in Appendix~\ref{app:analyses}, captures the hierarchy (.98 native, .82
BR).
Overall, SPF is high in the three scenarios in which the fitted forest uses
common conditions for both labels (D1, D2, D4) and zero in the two in which it
does not (D3, D5), irrespective of whether the labels share features.

\subsection{Native multi-output forests versus Binary Relevance (RQ4)}
\label{sec:res-real}

\begin{figure}[!htbp]
\centering
\includegraphics[width=\textwidth]{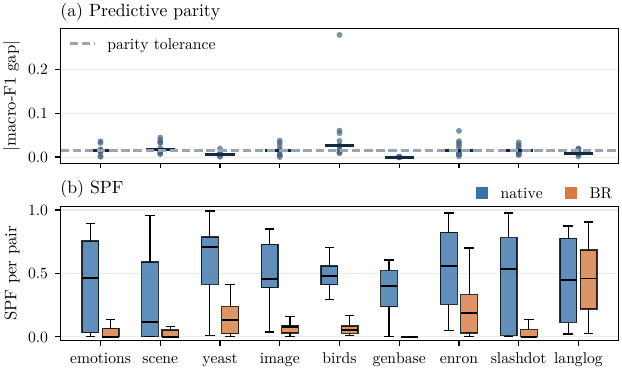}
\caption{Native and BR forests on the nine real benchmarks. (a) Absolute macro-F1 gap;
dashed line: .015 parity tolerance. (b) Median SPF over ordered pairs.}
\label{fig:real-native-br}
\end{figure}

Native multi-output forests share more decision structure than BR forests on eight of the nine benchmarks (Table~\ref{tab:real} and Fig.~\ref{fig:real-native-br}). On these eight, the median SPF excess (native $-$ BR) ranges from .107 (\emph{scene}) to .448 (\emph{slashdot}) and is positive in every repetition (one-sided Wilcoxon $p=.000977$, the smallest value attainable with ten repetitions; Holm-adjusted over the nine benchmarks, $p=.0088$). Six benchmarks meet the parity tolerance, and five of them show a positive excess. The three outside it (\emph{scene}, \emph{birds}, \emph{enron}) are reported to describe the full set of benchmarks, but are not evidence for the comparison at predictive parity. On \emph{genbase} both forests reach a test macro-F1 of 1 in almost every repetition, so that parity carries little information there; on \emph{birds}, with 129 test instances, the gap reaches .28 in one repetition.

\begin{table}[!htbp]
\caption{SPF difference native $-$ BR (32 quantile bins): median over repetitions of the median over ordered pairs, with IQR. Pos.: repetitions with a positive excess; Equal HP: excess at the BR hyperparameters; Rand.: absolute SPF difference between two such forests that differ only in their randomization; Par.: repetitions within the parity tolerance; $^\dagger$: median F1 gap above .015.}
\centering
\small
\setlength{\tabcolsep}{4pt}
\begin{tabular}{lrrrrrrr}
\toprule
Dataset & SPF excess [IQR] & Pos. & Equal HP & Rand. & Holm $p$ & F1 gap & Par.\\
\midrule
emotions & .419 [.338, .462] & 10 & .207 & .016 & .0088 & .0142 & 6\\
scene    & .107 [.085, .129] & 10 & .101 & .019 & .0088 & .0179$^\dagger$ & 4\\
yeast    & .426 [.409, .467] & 10 & .281 & .044 & .0088 & .0058 & 9\\
image    & .376 [.348, .396] & 10 & .298 & .047 & .0088 & .0140 & 6\\
birds    & .421 [.383, .488] & 10 & .456 & .053 & .0088 & .0261$^\dagger$ & 3\\
genbase  & .377 [.326, .390] & 10 & .377 & .039 & .0088 & .0000 & 10\\
enron    & .285 [.231, .331] & 10 & .104 & .043 & .0088 & .0160$^\dagger$ & 5\\
slashdot & .448 [.381, .501] & 10 & .754 & .002 & .0088 & .0142 & 5\\
langlog  & $-.032$ [$-.046$, $-.025$] & 2 & $-.046$ & .042 & .968 & .0084 & 8\\
\bottomrule
\end{tabular}
\label{tab:real}
\end{table}

\begin{figure}[!htbp]
\centering
\includegraphics[width=\textwidth]{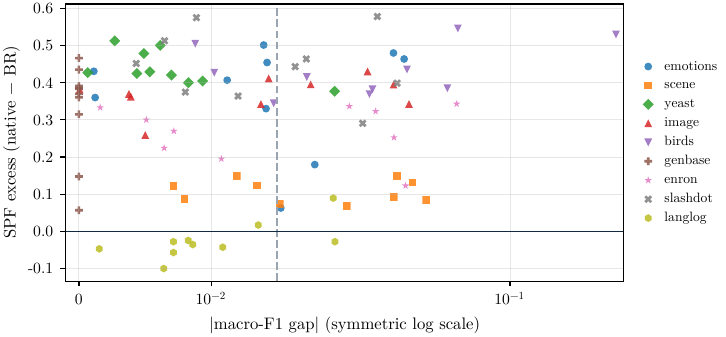}
\caption{SPF excess (native $-$ BR, median over ordered pairs) against the absolute test macro-F1 gap, one point per repetition. Dashed line: parity tolerance of .015.}
\label{fig:parity-gap}
\end{figure}

Because parity is assessed on the median gap, not every repetition meets it (column Par.). The excess does not follow the gap, however (Fig.~\ref{fig:parity-gap}): within each benchmark, the Spearman correlation between the two over the ten repetitions has a median of .05 (undefined on \emph{genbase}, where the gap is zero). Restricted to the repetitions within the tolerance, the median excess changes by at most .02 and remains positive on every benchmark except \emph{langlog}. This restriction selects repetitions on test macro-F1 and is a description, not a model-selection procedure.

\emph{langlog} is the exception. Its excess is positive in two of the ten repetitions (one-sided $p=.968$). It is also the benchmark with the lowest predictive performance for both architectures: their test macro-F1 is .058 and .044, its eight retained labels have a mean prevalence of .067, and half of its instances carry none of them. On \emph{langlog}, and on no other benchmark, the independently fitted BR forests already share much of their decision structure: their median SPF is .462, against at most .18 on the other eight.

Because the BR hyperparameters are tuned to match native macro-F1, this
comparison changes architecture and hyperparameters at the same time, and a BR model has $L$ times as many trees (Section~\ref{sec:experiments}). Because its trees are shallower, the total number of nodes is nevertheless of the same order: within a factor of 1.7 of the native forest on seven benchmarks, 3.5 times larger on \emph{genbase}, and 5.4 times smaller on \emph{slashdot}, whose native trees have a mean depth of 236 levels (Appendix~\ref{app:analyses}). Fitted with the BR hyperparameters (column Equal HP), the native forest still exceeds BR in every repetition of the eight benchmarks, by more than twice the difference between two such forests that differ only in their randomization (column Rand.). On \emph{emotions}, \emph{scene}, \emph{yeast}, \emph{image} and \emph{enron}, joint fitting accounts for 49\%, 94\%, 66\%, 79\% and 36\% of the raw excess, and the shallower trees and larger leaves selected
for BR account for the remainder. On \emph{genbase}, the grid selects the default hyperparameters for BR in every repetition, so that the two comparisons coincide. On \emph{birds} and \emph{slashdot}, the excess at equal hyperparameters is larger than the raw excess (.754 against .448 on \emph{slashdot}, where BR is limited to 8--16 levels in nine of the ten repetitions): the BR hyperparameters raise the sharing of the native forest, and the raw comparison understates the effect of joint fitting. On \emph{langlog} the excess stays negative at equal hyperparameters, in every repetition.

The smaller excess on \emph{scene} reflects its label structure: 93\% of its
instances carry a single label, an ordered pair shares a median of 9 active
predicates in the native forest (27.5 on \emph{emotions}, 108 on \emph{yeast}),
and 20\% of the pairs share none.

Table~\ref{tab:sens} reports two sensitivity analyses on the label set: \emph{yeast} with all 14 labels and, for the other benchmarks with more than eight labels, every label with at least 20 positive instances. The excess remains positive, also at equal hyperparameters, on all but \emph{genbase}, whose labels are nearly exclusive (90\% of the instances carry exactly one of the eight most prevalent labels): once its rarer labels are included, the median SPF over pairs is zero for both architectures. On \emph{langlog} the excess becomes positive but small, so the sign of its result depends on the labels retained.

\begin{table}[!htbp]
\caption{Sensitivity to the label set: \emph{yeast} with all labels, and the other benchmarks with more than eight labels with every label that has at least 20 positive instances. Columns as in Table~\ref{tab:real}; $p$: one-sided Wilcoxon test, not adjusted.}
\centering
\small

\begin{tabular}{lrrrrrr}
\toprule
Dataset & $L$ & SPF excess & Pos. & Equal HP & $p$ & F1 gap\\
\midrule
yeast    & 14 & .465 & 10 & .374 & .001 & .0043\\
birds    & 12 & .374 & 10 & .375 & .001 & .0296$^\dagger$\\
genbase  & 13 & .000 & 4  & .000 & .0625 & .0046\\
enron    & 35 & .264 & 10 & .168 & .001 & .0097\\
slashdot & 16 & .129 & 10 & .253 & .001 & .0136\\
langlog  & 25 & .028 & 9  & .010 & .0049 & .0026\\
\bottomrule
\end{tabular}

\label{tab:sens}
\end{table}

\subsection{SPF under label-permutation nulls}
\label{sec:res-arch}

Independent permutation of each training-label column preserves label
prevalence but removes two things at once: the dependence between labels and
the dependence of each label on the features. Two further nulls separate them. A joint permutation shuffles the rows of $Y$ as a block, which keeps
label dependence and removes the signal; a conditional permutation shuffles
each label within deciles of its out-of-fold score from a single-label forest,
which keeps most of the signal and removes the label dependence the features do not explain (Table~\ref{tab:nulls}, test AUC and $|\phi|$).

\begin{figure}[!htbp]
\centering
\includegraphics[width=\textwidth]{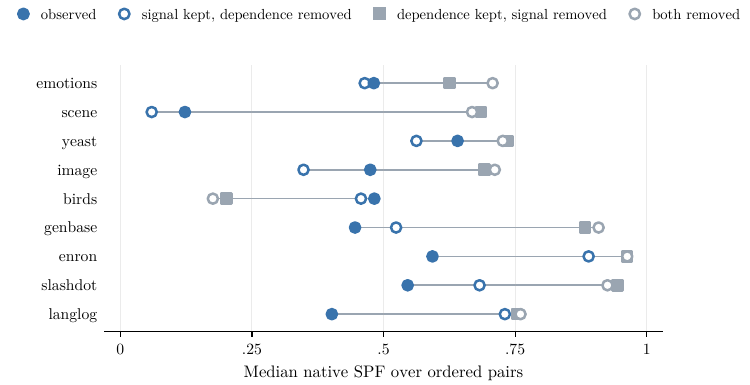}
\caption{Median native SPF with observed labels and under three label nulls,
by what each removes, on the nine benchmarks.}
\label{fig:permutation}
\end{figure}

Under independent permutation, native SPF increases on eight of the nine benchmarks (Fig.~\ref{fig:permutation}), for instance from .123 to .668 on \emph{scene}, and joint permutation gives a value within .03 of it on all of them except \emph{emotions} (.08). The increase is therefore due mainly to the loss of signal, not of label
dependence: labels without signal are still routed through the same splits, chosen to fit
the noise averaged over all outputs, whereas labels with distinct signal lead to different splits; the
increase is largest on \emph{scene}, whose labels are nearly exclusive.
\emph{birds} is the exception: there, removing the signal lowers native SPF from .483 to .176, so that this mechanism is common but not general.
Removing label dependence while keeping the signal has no consistent effect. It lowers SPF on \emph{emotions}, \emph{scene}, \emph{yeast}, \emph{image} and \emph{birds}, most on \emph{scene} (.123 to .060), and raises it on \emph{genbase}, \emph{enron}, \emph{slashdot} and \emph{langlog}, most on \emph{langlog} (.402 to .731). Since this null also removes part of the signal, lowering test AUC by .03 to .12 on every benchmark except \emph{genbase}, its effect cannot be attributed to label dependence alone. Raw SPF is therefore informative when comparing models
fitted to the same labels, not on an absolute scale.

\begin{table}[!htbp]
\caption{Native forest under label nulls: median SPF over ordered pairs, test AUC and mean $|\phi|$ (medians over repetitions). Cond.: dependence removed, signal kept; Joint: dependence kept, signal removed; Ind.: both removed.}
\centering
\footnotesize

\begin{tabular}{lrrrrrrrr}
\toprule
 & \multicolumn{4}{c}{SPF} & \multicolumn{2}{c}{Test AUC} & \multicolumn{2}{c}{$|\phi|$}\\
\cmidrule(lr){2-5}\cmidrule(lr){6-7}\cmidrule(lr){8-9}
Dataset & Obs. & Cond. & Joint & Ind. & Obs. & Cond. & Obs. & Cond.\\
\midrule
emotions & .482 & .464 & .625 & .707 & .850 & .814 & .325 & .203\\
scene    & .123 & .060 & .685 & .668 & .951 & .925 & .185 & .108\\
yeast    & .641 & .563 & .735 & .727 & .751 & .695 & .241 & .063\\
image    & .475 & .348 & .692 & .711 & .838 & .778 & .220 & .059\\
birds    & .483 & .457 & .201 & .176 & .896 & .794 & .078 & .073\\
genbase  & .446 & .524 & .883 & .909 & 1.000 & .997 & .122 & .094\\
enron    & .593 & .890 & .962 & .963 & .837 & .764 & .145 & .074\\
slashdot & .546 & .682 & .944 & .925 & .872 & .788 & .094 & .036\\
langlog  & .402 & .731 & .753 & .760 & .784 & .669 & .044 & .030\\
\bottomrule
\end{tabular}

\label{tab:nulls}
\end{table}

\subsection{Sensitivity to predicate matching and activity cut-offs}
\label{sec:res-robust}

On the synthetic scenarios of Section~\ref{sec:synth}, with top-percentile thresholds of 10--50\% and minimum-gain thresholds of 1--10\%, SPF remains exactly zero in the two scenarios without shared conditions, its medians remain at or above .47 in the three with shared conditions, and the ordering of the scenarios is preserved (Kendall's $\tau\geq.74$ relative to the default).

\begin{figure}[!htbp]
\centering
\includegraphics[width=\textwidth]{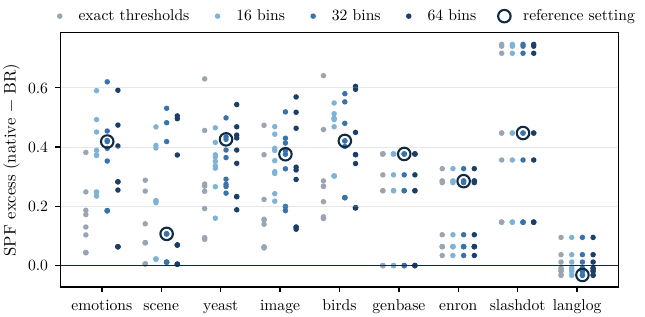}
\caption{SPF excess (native $-$ BR) of each benchmark under the 36 extraction settings, coloured by matching rule; circles mark the reference setting.}
\label{fig:robustness}
\end{figure}

On the real benchmarks, SPF was computed for the native and BR forests of Section~\ref{sec:res-real}, in every repetition, under 36 predefined settings: four matching rules (exact thresholds, and 16, 32 or 64 quantile bins) crossed with the nine combinations of the activity cut-offs (top 10, 25 or 50\% of the predicates with positive gain, and a minimum of 1, 5 or 10\% of the maximum gain). Fig.~\ref{fig:robustness} shows, for each benchmark, the median excess under each setting, one point per setting, coloured by matching rule, with a circle at the reference setting used in the rest of the paper (32 quantile bins, top 25\%, minimum 5\%). A benchmark whose points all lie above zero has a positive excess whatever the setting, and the vertical spread of its points shows how much the magnitude of the excess depends on the setting. Appendix~\ref{app:analyses} gives further summaries.

The sign of the excess is stable. It is positive under all 36 settings on the seven benchmarks other than \emph{genbase} and \emph{langlog}. On \emph{genbase} it vanishes under the 10\% minimum, which leaves each label with about two active predicates. On \emph{langlog} it remains negative under the 5\% and 10\% minima and becomes positive (.01 to .09) under the 1\% minimum. The magnitude is not stable: it ranges, for instance, from .004 to .53 on \emph{scene} and from .03 to .33 on \emph{enron}, and is smallest with exact matching and with the 10\% minimum, the settings that leave the fewest predicates to share. On \emph{genbase}, \emph{enron}, \emph{slashdot} and \emph{langlog}, whose features are binary, every matching rule yields the same predicates, and only the activity cut-offs matter. Under the 27 settings that match by quantile bins, the ranking of label pairs by excess has a median Spearman correlation with the ranking under the reference setting between .48 (\emph{langlog}) and .93 (\emph{scene}), and close to zero on \emph{genbase}. Comparisons of SPF values are therefore meaningful under a fixed setting, as in the rest of this section, and the magnitudes of Table~\ref{tab:real} are specific to the reference setting.

An unweighted comparison of the active sets shows a much smaller difference between architectures. The Jaccard index of the two labels' active sets is only slightly higher in the native forest than under BR: its median excess at the reference setting is between .006 and .106 on the eight benchmarks with a positive SPF excess, against SPF excesses between .107 and .448, and $-.016$ on \emph{langlog}. It is positive under all 36 settings on six benchmarks, under 33 on \emph{scene}, 24 on \emph{genbase} and 8 on \emph{langlog}. Within the native forest, SPF and the Jaccard index rank label pairs similarly (median Spearman correlation between .54 and .96 at the reference setting). The SPF excess therefore reflects how much of each label's decision mass passes through shared predicates, more than a larger overlap of the active sets. The native active sets are nevertheless larger on every benchmark (for example 340 against 96 predicates per label on \emph{emotions}), consistent with Lemma~\ref{lem:gini}, by which a split credits gain to every label whose child rates differ.

\subsection{Feature attributions under a change of architecture}
\label{sec:res-shap}

Table~\ref{tab:shap} compares the per-label mean $|$TreeSHAP$|$ rankings of the forests of Section~\ref{sec:res-real}. Between a native forest fitted with the BR hyperparameters and BR itself, the median Spearman agreement is lower than between two such native forests that differ only in their randomization on all nine benchmarks, by .01 on \emph{genbase} and by .47 on \emph{slashdot}; changing the hyperparameters alone leaves the rankings almost as close as a change of randomization. On \emph{slashdot} the comparisons with the default native forests are missing, since their trees are too deep for TreeSHAP (Section~\ref{sec:baselines}), and the others rest on nine repetitions; on \emph{enron} the hyperparameter-only comparison uses the 37 of 80 label rankings of the default native forests that satisfy the additivity criterion.

\begin{table}[!htbp]
\caption{Spearman agreement of per-label mean $|$TreeSHAP$|$ rankings (median over labels and repetitions). Architecture: native forest at the BR hyperparameters against BR; Randomization: two such native forests; Hyperparameters: default against BR-hyperparameter native forest; --: not computable.}
\centering
\small

\begin{tabular}{lrrr}
\toprule
Dataset & Architecture & Randomization & Hyperparameters\\
\midrule
emotions & .71 & .93 & .90\\
scene    & .69 & .90 & .93\\
yeast    & .71 & .93 & .94\\
image    & .80 & .92 & .94\\
birds    & .81 & .92 & .92\\
genbase  & .96 & .97 & 1.00\\
enron    & .78 & .88 & .86\\
slashdot & .36 & .83 & --\\
langlog  & .64 & .75 & .85\\
\bottomrule
\end{tabular}

\label{tab:shap}
\end{table}

\subsection{Case study: a label pair in \emph{emotions}}
\label{sec:res-emotions}

\begin{figure}[!htbp]
\centering
\includegraphics{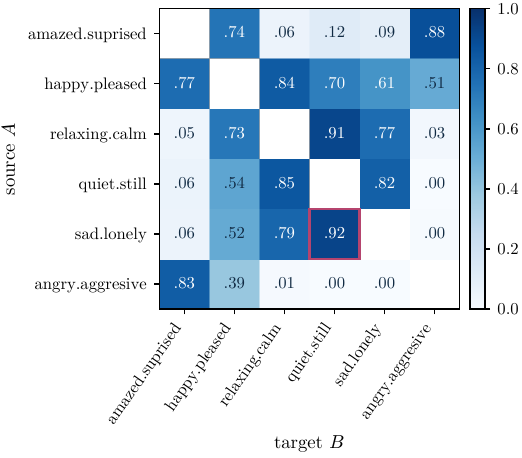}
\caption{SPF matrix of the native forest on \emph{emotions} (first repetition); the pair of Fig.~\ref{fig:msdpg-graph} is outlined.}
\label{fig:spf-emotions}
\end{figure}

The SPF matrix of the native forest fitted to \emph{emotions} in the first repetition (Fig.~\ref{fig:spf-emotions}) shows two groups of labels decided through common conditions: \texttt{relaxing.calm}, \texttt{quiet.still} and \texttt{sad.lonely}, with SPF between .77 and .92 among them, and \texttt{amazed.suprised} and \texttt{angry.aggresive} (.83 and .88), with SPF of at most .12 between the two groups; \texttt{happy.pleased} shares conditions with most labels.

\begin{sidewaysfigure}
\centering
\includegraphics[width=\textheight]{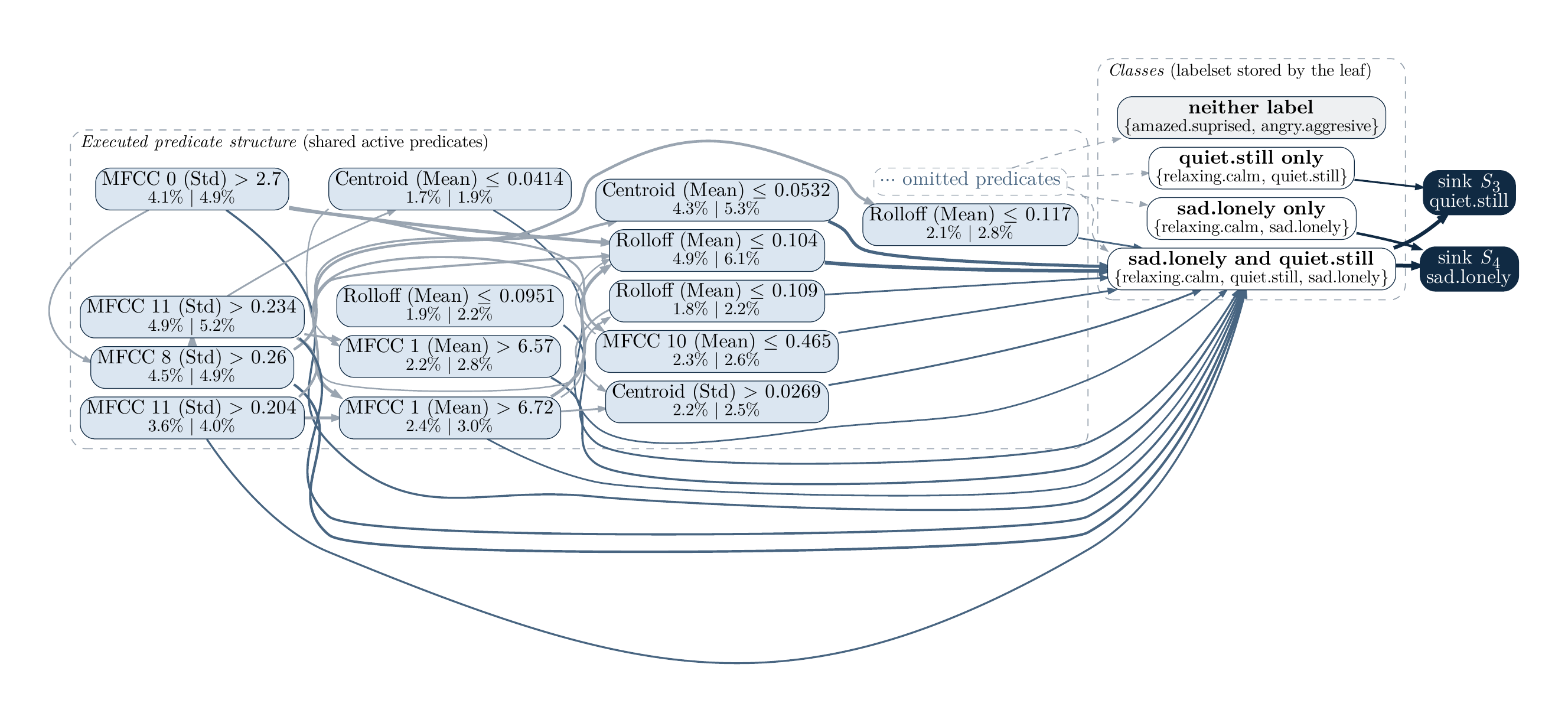}
\caption{Contracted ms-DPG for \texttt{sad.lonely}$\to$\texttt{quiet.still} (\emph{emotions}, first repetition): shared active predicates ordered by mean depth, with their share of each sink's mass; grey arcs are executed transitions, possibly through omitted predicates (dashed node); leaves are grouped by the labelset they store for the pair. Edge width is proportional to mass.}
\label{fig:msdpg-graph}
\end{sidewaysfigure}

In this forest, the ordered pair with
the highest SPF links \texttt{sad.lonely} to \texttt{quiet.still}
(Fig.~\ref{fig:msdpg-graph}). Of its 131 shared active predicates, 130 are
co-enriching (Fig.~\ref{fig:lift}a), and the fourteen displayed describe how
the forest decides the two labels: from low mean spectral rolloff and centroid,
which characterize excerpts whose spectral energy is concentrated at low
frequencies and whose timbre is darker~\cite{tzanetakis2002genre}, from high
mean MFCC~1 and low mean MFCC~10, and, closer to the root of the trees, from
high variability of MFCC~0, MFCC~8 and MFCC~11. The graph shows how these
conditions are chained along the executed paths, from the MFCC variability
tested near the root to rolloff and centroid tested deeper, and that the
forest tests several neighbouring thresholds of the same condition, for
instance mean rolloff at $0.095$, $0.104$, $0.109$ and $0.117$. Of the full graph (4,516 predicates, 48,774 transitions, six sinks), the view shows 14 of the 131 shared active predicates. Its leaves, all pure, are grouped into four classes by the labelset they store for the pair, describing tree-level outputs rather than ensemble predictions; the classes hold 17.9\% (both), 10.6\% (\texttt{sad.lonely} only), 6.8\% (\texttt{quiet.still} only) and 64.7\% (neither) of the training mass, and each predicate is linked to the class that receives most of its mass. The class
layer adds that the leaves storing both emotions, class \emph{both}, most often
store a full labelset that also contains \texttt{relaxing.calm} (69\% of that
class's mass). Among the classes that reach a sink, every displayed predicate
sends more of its mass to class \emph{both} than to any other class, and for ten of the
fourteen, all but the four variability conditions near the root, this class
receives most of the mass outright (52--79\%). Each class node names the
labelset most frequent within the class: \{relaxing.calm, quiet.still,
sad.lonely\} accounts for 69\% of class \emph{both},
\{relaxing.calm, sad.lonely\} for 33\% of \texttt{sad.lonely} only,
\{relaxing.calm, quiet.still\} for 57\% of \texttt{quiet.still} only, and
\{amazed.suprised, angry.aggresive\} for 22\% of class \emph{neither}.
The displayed predicates cover class \emph{both} best: 51\% of
its mass lies on paths through none of them, against 84\%, 71\% and 89\% for
the other three classes.

\subsection{Case study: a label pair in \emph{flags}}
\label{sec:res-flags}

\begin{sidewaysfigure}
\centering
\includegraphics[width=\textheight]{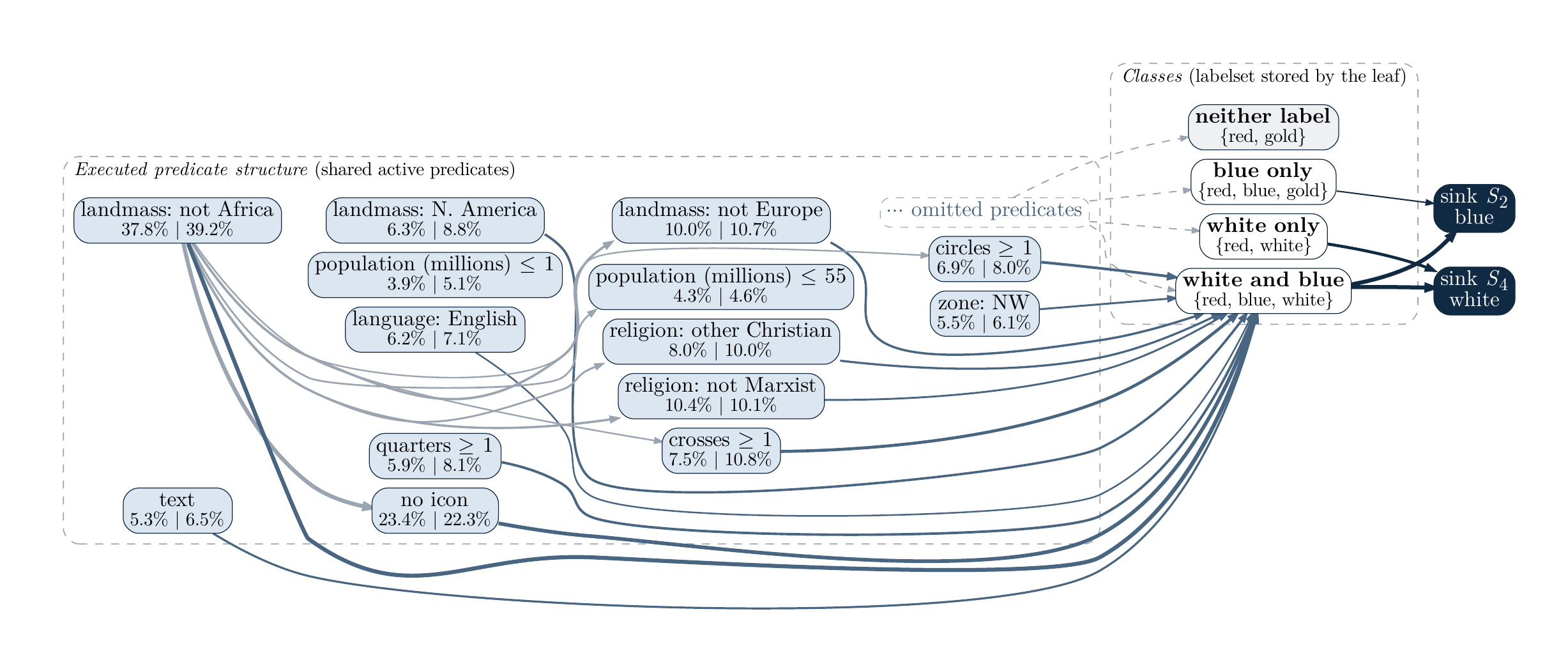}
\caption{Contracted ms-DPG for \texttt{white}$\to$\texttt{blue} (\emph{flags}, first repetition), conventions as in Fig.~\ref{fig:msdpg-graph}.}
\label{fig:msdpg-flags}
\end{sidewaysfigure}

On \emph{flags} (114 training flags, test macro-F1 .58), SPF is close to
saturation, several ordered pairs reaching about .96, so the case is read
qualitatively. For the highest-SPF pair,
\texttt{white}$\to$\texttt{blue} (SPF .97), 17 of the 31 shared active
predicates are co-enriching (Fig.~\ref{fig:msdpg-flags}). The forest raises the probability of both colours for flags with a quartered canton (lift 1.34 for
\texttt{white} and 1.84 for \texttt{blue}), a cross (1.25 and 1.80) or a circle
(1.21 and 1.39), and for countries of other Christian religion (1.10 and
1.37): among flags with a cross, for example, the forest predicts blue 1.8
times and white 1.25 times as often as on average, consistent with the Nordic
crosses and the flags derived from the Union Jack. These statements describe a model fitted to 114 flags with modest accuracy; they illustrate how the explanation reads, not established regularities about national flags. Not all sharing is joint,
however. Separating conditions carry 33\% of the mass of the shared
predicates; for flags with two or more stripes, for instance, the forest
predicts blue more often (1.14) and white less often (.89) than on average. The
explanation thus shows that the model raises the probability of white and blue largely through the same cross and canton designs, while among striped flags it tends to
predict one colour rather than the other. The view shows 14 of the 31 shared active predicates, out of 204 predicates in the graph. Class \emph{both} holds 44.7\% of the leaf mass, the classes \texttt{white} only, \texttt{blue} only and neither 29.6\%, 7.3\% and 18.3\%, and its
most frequent labelset, \{red, blue, white\}, accounts for 34\% of it; the most
frequent labelsets of the other classes are \{red, white\} (21\% of
\texttt{white} only), \{red, blue, gold\} (38\% of \texttt{blue} only) and
\{red, gold\} (24\% of neither). Paths through none of the displayed predicates
carry 16\% of the mass of class \emph{both}, against 34\%,
36\% and 42\% for the other classes.

\paragraph{Coverage and stability of the views.} Each contracted view displays 14 predicates and omits the others. In the views of Figs.~\ref{fig:msdpg-graph} and~\ref{fig:msdpg-flags}, the traces through the displayed predicates carry 37\% and 44\% of the decision mass of \texttt{sad.lonely} and \texttt{quiet.still}, and 77\% and 82\% of that of \texttt{white} and \texttt{blue}; over the ten repetitions, the medians are 44\% and 52\% on \emph{emotions} and 79\% and 82\% on \emph{flags}. The omitted mass is largest in the class of leaves that store neither label (median 88\% on \emph{emotions} and 43\% on \emph{flags}) and smallest in the class that stores both (42\% and 16\%). The selection itself varies across repetitions. Matching predicates of different repetitions when they test the same feature in the same direction with overlapping threshold bins, the median Jaccard index between the displayed sets of two repetitions is .27 on \emph{emotions} and .33 on \emph{flags}. Of the 14 predicates of Fig.~\ref{fig:msdpg-graph}, four are displayed again in at least five of the other nine repetitions, and nine are shared and active in at least five; for Fig.~\ref{fig:msdpg-flags} the counts are six and eight. Whether a shared predicate is co-enriching or separating is stable on \emph{emotions}, where the classification agrees across repetitions in 99.7\% of the comparisons, weighted by mass, and less so on \emph{flags} (65\%), whose forests are fitted to 114 instances. A contracted view therefore describes the forest of one repetition, and the predicates that recur across repetitions are the most reliable part of its reading (Appendix~\ref{app:analyses}).

\section{Discussion}
\label{sec:discussion}

\subsection{A global, model-specific and structural explanation}

ms-DPG provides a global, model-specific and structural explanation. Its unit
is a directional decision condition, neither a feature nor an instance, and
the question it answers concerns a pair of outputs: whether the fitted forest
decides two labels through the same conditions, which conditions these are,
and whether the regions they delimit are enriched in both labels or separate them. Because its sinks are
the labels, cross-label reuse is read from a single graph instead of being
inferred from a comparison of per-label rankings (Table~\ref{tab:related}).

The explanation is true to the model in the sense of Chen et
al.~\cite{chen2020true}, which accounts for the D2 and permutation results.
SPF is high in D2 because the fitted forest uses the correlated proxy $x_1$
for label~1, and it increases when label permutation removes the feature--label signal (on eight of the nine benchmarks),
because a native forest routes every label through the same splits. Raw SPF thus combines the sharing
imposed by the architecture with the sharing induced by the labels, and the
pair-level view is needed to separate the two.

\subsection{Complementarity with feature attribution}
\label{sec:disc-attr}

Per-label attributions may suggest that two labels rest on the same evidence
when the forest decides them through opposite conditions on it, as in D5. They
also register a change of architecture, but not its
nature: across the nine benchmarks, the per-label TreeSHAP rankings of native and BR forests with the same hyperparameters agree less than those of two forests that differ only in their randomization (Section~\ref{sec:res-shap}), most on \emph{slashdot}. Even on \emph{langlog}, where native forests share no more than BR, the rankings change: attributions register that the architecture differs, but not whether it shares more. An analyst comparing
these rankings would observe reshuffled features, moderately on most benchmarks, while the ms-DPG
locates the change in the larger share of each label's mass that the native
model routes through conditions shared with other labels
(Table~\ref{tab:real}). The choice between native and BR forests is therefore
not neutral for explanation, because in the native model the conditions that
decide one label also decide others.

\begin{figure}[!htb]
\centering
\includegraphics[width=\textwidth]{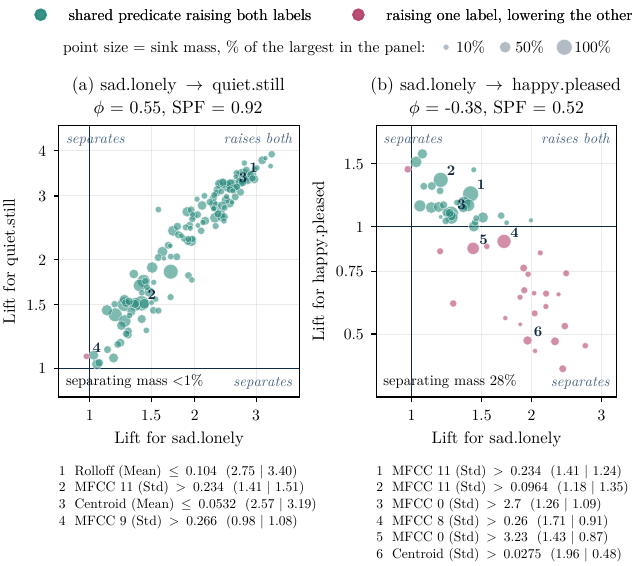}
\caption{Lifts of the shared active predicates of two \emph{emotions} pairs
(first repetition): co-enriching (green) and separating (magenta); point size is decision
mass. Numbered predicates are listed below with their lifts (source | target).}
\label{fig:lift}
\end{figure}

The lift view adds the direction of sharing, which SPF does not encode
(Fig.~\ref{fig:lift}). A lift describes what the forest predicts in the region a
condition delimits: among the training mass that passes through it, the trees assign the
label on average $\lambda$ times the probability they assign it overall (below,
``predicts $\lambda$ times as often''). For excerpts with low
mean spectral rolloff ($\leq .104$), for instance, the forest predicts
\texttt{sad.lonely} 2.75 times and \texttt{quiet.still} 3.40 times as often as
on average, so that the forest raises the probability of both emotions where this condition holds. For
\texttt{sad.lonely} and \texttt{quiet.still} ($\phi=.55$, SPF $=.92$), all
shared predicates but one, which carries less than 1\% of their mass, are
co-enriching in this sense, and the forest largely raises the two emotions through the same conditions. For \texttt{sad.lonely} and \texttt{happy.pleased} ($\phi=-.38$)
SPF is still .52, yet 28\% of the combined decision mass of the shared predicates
lies on separating conditions. Under high centroid variability ($>.028$) the
forest predicts \emph{sad} 1.96 times as often and \emph{happy} only .48
times as often, and high mean MFCC~1 ($>5.52$) has lifts of 2.29 and .48: the
two emotions share these conditions, but the forest uses them to choose one
over the other. Here the local reading agrees: at the splits that produce these
conditions, the rate of \emph{happy} falls relative to the parent node (.44
and .72) while that of \emph{sad} rises (1.42 and 1.33). SPF alone would report substantial sharing for this pair
without revealing that part of it is opposition. The atlas of Appendix~\ref{app:atlas} shows the same on \emph{enron}, where the pair with the highest SPF (.99) has 94\% of its shared mass on separating predicates, and, on \emph{yeast}, a pair with SPF .99 whose shared predicates have lifts close to 1, that is, sharing that the lifts do not differentiate. Across all ordered pairs, the
separating share of shared mass decreases as label association increases
(Spearman $-.75$, $-.27$ and $-.56$).

Read together, SPF and the lifts answer questions about the fitted model that
per-label explanations leave open. Whether the forest tends to predict
\texttt{quiet.still} when it predicts \texttt{sad.lonely} is answered by a high
SPF with co-enriching conditions; whether it treats \texttt{sad.lonely} and
\texttt{happy.pleased} as opposites is answered, in part, by the separating
conditions it uses for them; and, for any condition, the lifts indicate which
other labels the forest decides through it and in which direction. These readings describe how the model decides, not how the world
behaves: a lift is neither a property of the data nor a causal effect of the
feature.

\subsection{Purposes of explainability}

Of the purposes of explanation distinguished by Adadi and
Berrada~\cite{adadi2018peeking}, ms-DPG serves three: explaining to control,
by revealing that outputs expected to be decided independently share
conditions, and in which direction; explaining to improve, since SPF informs
the choice between native and BR forests at comparable accuracy; and
explaining to discover, since shared predicates can reveal a proxy for a
label's driver (D2) or labels decided by the same region of the input space.

\paragraph{Intended users and workflow.} The explanation is addressed to those who build, select or audit a multi-output model, rather than to the recipients of its predictions. Its use follows four steps, illustrated on \emph{emotions} in Section~\ref{sec:res-emotions} and on all nine benchmarks in Appendix~\ref{app:atlas}. First, the SPF matrix of the fitted forest (Fig.~\ref{fig:spf-emotions}) identifies the label pairs whose decision mass runs through common conditions; when BR is an alternative, the excess over BR shows how much of this sharing is due to joint fitting. Second, for a selected pair, the contracted view names the shared predicates and the order in which the trees test them (Fig.~\ref{fig:msdpg-graph}). Third, the lifts state whether each shared predicate raises the probability of both labels or of one at the expense of the other (Fig.~\ref{fig:lift}), which distinguishes labels that the model raises together from labels that it treats as alternatives. Fourth, the predicates that recur across fits on different data partitions are retained as the stable part of the reading (Section~\ref{sec:res-flags}).

\paragraph{Evaluation against explanation properties.} Nauta et al.~\cite{nauta2023anecdotal} group the properties on which explanations are evaluated into twelve, the Co-12 properties. Table~\ref{tab:co12} relates the evidence of this paper to them. Correctness and completeness hold analytically for the full graph, which reproduces the traces, masses and probabilities of the forest (Proposition~\ref{prop:mass}); the contracted views trade completeness for compactness, and cover 37\% to 82\% of the decision mass of the two labels in the case studies. Consistency has two parts that the evidence treats separately: the extraction is deterministic for a given model, whereas the predicates displayed for a pair vary moderately across refitted models. Continuity in the sense of the taxonomy, the response of the explanation to small changes of the input, was not assessed; what was assessed is the sensitivity to the extraction settings, under which the sign of the native--BR comparison is stable and its magnitude is not. The table therefore collects evidence bearing on each property, not a validation of it. Contrastivity is the property on which the explanation adds most to feature-level summaries: it separates co-enriching from separating conditions and responds to the direction of a condition, to which per-label feature summaries are blind (Propositions~\ref{prop:complement} and~\ref{prop:dissociation}). Coherence with domain knowledge is supported only by the two case studies, and confidence, context and controllability have not been evaluated; in particular, no study with users was conducted.

\begin{table}[!htbp]
\caption{Evidence bearing on the Co-12 properties of explanation quality~\cite{nauta2023anecdotal}; partial evidence, not a validation of each property.}
\centering
\scriptsize
\setlength{\tabcolsep}{3pt}
\renewcommand{\arraystretch}{1.12}
\begin{tabular}{@{}p{2.1cm}p{4.2cm}p{5.3cm}@{}}
\toprule
Property & Meaning for ms-DPG & Evidence\\
\midrule
Correctness & Graph and sinks reproduce the forest's traces, masses and probabilities & Proposition~\ref{prop:mass}; identity with the reference DPG implementation (Section~\ref{sec:experiments})\\
Completeness & Share of the decision mass that the explanation covers & Holds for the retained traces of the full graph; the contracted views cover 37--82\% (Section~\ref{sec:res-flags})\\
Consistency & (a) Same model, same explanation; (b) refitted models, similar explanations & (a) Deterministic extraction; (b) displayed predicates recur moderately across repetitions, lift classes agree in 65--99.7\% (Section~\ref{sec:res-flags})\\
Parameter sensitivity (in place of continuity) & Dependence on the extraction settings & Sign of the native--BR excess stable under 36 settings on seven of nine benchmarks, magnitude not (Section~\ref{sec:res-robust}); continuity with respect to the input was not assessed\\
Contrastivity & Distinguishes labels, directions and architectures & Scenario D5, Propositions~\ref{prop:complement}--\ref{prop:dissociation}; co-enriching against separating predicates; native against BR (Section~\ref{sec:res-real})\\
Covariate complexity & Interpretability of the explanatory unit & Directional predicates on the original features, readable as far as the features are (Appendix~\ref{app:atlas})\\
Compactness & Size of what is shown & $L(L-1)$ SPF values; 14 predicates per contracted view\\
Composition & Organization of the explanation & Two levels, SPF matrix and contracted view with lifts; reading guide (Table~\ref{tab:reading})\\
Coherence & Agreement with domain knowledge & Case studies only (Sections~\ref{sec:res-emotions}--\ref{sec:res-flags})\\
Confidence, context, controllability & Uncertainty, user needs, interaction & Not evaluated\\
\bottomrule
\end{tabular}
\label{tab:co12}
\end{table}

\section{Limitations}
\label{sec:limitations}

SPF does not measure statistical label dependence (Fig.~\ref{fig:permutation})
and follows correlated proxies (D2). It measures how much of the target label's mass
lies on traces with at least one predicate active for both labels, not how
much of each trace is shared (Section~\ref{sec:spf}). It describes which conditions the model
shares; the effect on one label of changing the inputs behind another is an
interventional question outside its scope. Comparisons across architectures
depend on the definition of predicate equivalence (Section~\ref{sec:res-matching}), on the activity cut-offs, which change the magnitude of the native--BR difference though rarely its sign (Section~\ref{sec:res-robust}),
and on hyperparameters, which account for part of the raw native--BR
difference. The \emph{scene}, \emph{birds} and \emph{enron} comparisons, finally, fall outside the macro-F1
parity tolerance, and on \emph{langlog}, where both architectures predict poorly, native forests do not share more decision structure than BR. The quadratic cost in label pairs and the number of sinks limit the method to settings with tens of labels; thousands of labels are outside the scope of this work. The quantitative comparison rests on nine benchmarks with at most 35 labels, taken from those available on OpenML, so the magnitudes reported here may not carry over to other domains, label densities or larger label spaces. The TreeSHAP comparisons are incomplete on \emph{enron} and \emph{slashdot}, whose deep native trees make path-dependent TreeSHAP numerically unstable. All masses, and hence SPF and the lifts, are computed on the in-bag training sample; whether the shared structure carries the same weight on unseen data is not measured. The contracted views display a selection of predicates and can omit a substantial part of the decision mass; their selection varies across repetitions (Section~\ref{sec:res-flags}), and their usefulness to analysts has not been evaluated. The analytical results comparing native and BR forests (Section~\ref{sec:theory}) hold on the population and at limited depth, and do not quantify the excess observed in fully grown trees.

\section{Conclusion}

We introduced ms-DPG and Shared Predicate Flow to explain how a native
multi-output tree ensemble shares directional decision conditions across its
outputs. On RQ1, how to represent decision structure shared across outputs while preserving label-specific information, the ms-DPG represents this shared structure as a single
predicate graph whose sinks are the labels, and its contracted view, with
label-specific lifts, names the conditions a pair of outputs shares and shows
whether, where they hold, the model assigns a higher probability to both labels or to one at the expense of the other.
On RQ2, whether this sharing can be quantified at the level of directional conditions, SPF quantifies the sharing for every ordered pair at the level of
directional conditions. On the synthetic scenarios it is robust to the activity cut-offs when predicates are matched by threshold bin; on the real benchmarks, the sign of the native--BR difference is robust to the matching rule and the cut-offs on seven of the nine benchmarks, whereas its magnitude depends on them. On RQ3, whether predicate-level explanation reveals structure that label association and per-label relevance cannot,
SPF responds to shared decision conditions and not to shared features or label
association, as the disjoint-region scenario shows, whereas per-label
attributions register a change of architecture only as a reshuffling
of features, which does not show whether the model shares more. On RQ4, how joint fitting changes this structure relative to independently fitted label models, native forests share substantially more decision
structure than Binary Relevance on eight of nine benchmarks, including five of the six on which the two architectures are at predictive parity, largely because of
the architecture itself, which routes all labels through the same splits even when the labels carry no signal from the features. A
signed, effect-based variant of SPF that distinguishes co-enriching from
separating conditions within the measure itself is a natural extension.

The formal results of Section~\ref{sec:theory} support these answers: complementing a label leaves per-label summaries unchanged while mirroring its active predicates, and the joint criterion can select a shared condition that is not the best for any single label. Further work should extend this analysis to fully grown trees and finite samples, extend the quantitative evaluation to larger label spaces and to settings of low predictive performance, such as \emph{langlog}, where the excess was not observed, and apply ms-DPG to other multi-output tree learners, such as gradient-boosted ensembles and predictive clustering trees. Because ms-DPG describes the fitted model over the training distribution, an out-of-sample variant that weights traces by held-out instances would show whether the shared structure persists on unseen data. Finally, the selection of the contracted view varies across repetitions, which motivates a selection rule that favours predicates recurring across fits, and its usefulness to human analysts calls for dedicated studies.

\appendix
\section{Supplementary Formal Results}
\label{app:theory}

\paragraph{Dependence on depth.} Item (iv) of Proposition~\ref{prop:range} relates coverage to the number of shared predicates on a trace and to its length. Let $q=\sum_p m_ph_p/\sum_p m_p\ell_p$ be the mass-weighted density of shared predicates and $D=\sum_p m_p\ell_p/\sum_p m_p$ the mean trace length. If shared occurrences were independent along a trace with density $q$, SPF would be the mass-weighted mean of $1-(1-q)^{\ell_p}$, which by Jensen's inequality is at most $1-(1-q)^D$; we use the latter as a simple approximation. This is a heuristic, not a theorem, and since $q$ and $D$ are measured on the same forest it decomposes SPF rather than predicting it. Over \emph{emotions}, \emph{scene} and \emph{yeast}, ten repetitions, and observed and independently permuted labels (60 forests), the approximation is within .05 of the median SPF over ordered pairs in every forest, and its rank correlation with SPF across pairs is at least .90 in 59 of the 60 (minimum .835). It helps to read the permutation results of Section~\ref{sec:res-arch}: permuting the labels raises $D$ from 8.4 to 9.6, from 11.1 to 15.0 and from 12.1 to 14.9, and $q$ from .066 to .120, from .012 to .065 and from .082 to .083, on \emph{emotions}, \emph{scene} and \emph{yeast}. On \emph{yeast} the rise thus comes almost entirely from longer traces. By Lemma~\ref{lem:gini}, a split selected on noise still credits positive gain to every label whose child rates differ.

\begin{proposition}[Selection in scenario D1]
\label{prop:d1}
In scenario D1 with label noise $\varepsilon<\frac12$, consider the population criterion at a node reached only through features independent of $(x_0,x_1,x_2)$. Among splits on $x_0,x_1,x_2$, the native criterion is uniquely maximized by $x_0$ at threshold 0, with twice the summed gain of the best split on $x_1$ or $x_2$. For BR, the forest of $y_0$ is indifferent between $x_0$ and $x_1$ at threshold 0, and that of $y_1$ between $x_0$ and $x_2$.
\end{proposition}

Proposition~\ref{prop:d1} agrees with Section~\ref{sec:res-matching}: both architectures recover the planted rule, native trees because the condition is preferred among the planted features, BR trees because it remains available after the tied split.

\paragraph{Scope of Proposition~\ref{prop:selection}.} Its SPF statements stop at these depths deliberately. For larger depths, BR trees also reach $\sigma_s$, and SPF depends on the activity rule: with $\alpha=.2$, $\beta=.25$ and $m=1$, native and BR forests both have SPF 0 at depth 3, because with so few distinct predicates the top-25\% rule excludes the shared one. The proposition explains how the joint criterion can bring a shared condition forward; it does not account for the size of the excess observed in fully grown trees on the real benchmarks (Section~\ref{sec:res-real}). Results for finite samples and for random feature subsets are not covered.

\paragraph{Numerical checks.} Propositions~\ref{prop:mass}, \ref{prop:complement} and~\ref{prop:lift} hold to floating-point precision on forests fitted to \emph{emotions}, using the implementation of our experiments, and the bounds of Proposition~\ref{prop:range}(iv) hold in all 60 forests above. Both constructions of Proposition~\ref{prop:dissociation} return SPF 1 and 0. With 200,000 instances, stump forests under Proposition~\ref{prop:selection} give native SPF .699 for $(\alpha,\beta,m)=(.2,.25,1)$ and .601 for $(.1,.12,3)$, against predicted values of .70 and .60, and BR SPF 0. With $4\times10^6$ instances, the maximizers and gains of Proposition~\ref{prop:d1} match their theoretical values to within .0005.

\section{Proofs}
\label{app:proofs}

\begin{proof}[of Lemma~\ref{lem:gini}]
Let $p=P_Lp_L+P_Rp_R$ be the rate at the node. Since the binary Gini impurity is $2p(1-p)$ and $p$ is linear in the child rates, $\Delta G_\ell=2(P_Lp_L^2+P_Rp_R^2-p^2)=2P_LP_R(p_L-p_R)^2$, using $P_L+P_R=1$. Replacing $y_\ell$ by $1-y_\ell$ replaces $p_L-p_R$ by its negative. \qed
\end{proof}

\begin{proof}[of Proposition~\ref{prop:mass}]
(i) At a leaf $r$ of $T_k$, $w_r=\sum_{i:\,r_k(\mathbf{x}_i)=r}c_{ki}$ and $p_{r\ell}$ is the in-bag positive fraction, so $m(r,S_\ell)=\sum_{i:\,r_k(\mathbf{x}_i)=r}c_{ki}y_{i\ell}$. The leaves of a tree partition its in-bag instances; summing over leaves and trees gives the claim. (ii) Tree $k$ outputs $p_{r_k(\mathbf{x}),\ell}=m(r_k(\mathbf{x}),S_\ell)/w_{r_k(\mathbf{x})}$, and the forest averages these outputs. (iii) Each occurrence of a predicate on a trace has exactly one incoming and one outgoing arc, both carrying that trace's mass. Summing over all occurrences of the predicate, with multiplicity, gives equal inflow and outflow. \qed
\end{proof}

\begin{proof}[of Proposition~\ref{prop:range}]
Every $m_p$ with $p\in\mathcal{P}_B$ is positive and the indicator takes values in $\{0,1\}$, so the numerator of Eq.~\ref{eq:spf} is a sum over a subset of the terms of the denominator, which gives (i). The numerator vanishes exactly when all indicators are 0, and equals the denominator exactly when all are 1, which gives (ii) and (iii). For (iv), $h_p/\ell_p\leq\mathbb{1}_{A,B}(p)\leq h_p$ for every trace, since $h_p\leq\ell_p$ and $h_p$ is a non-negative integer that is positive exactly when the indicator is 1; for an empty trace all three quantities are 0. Multiplying by $m_p$, summing and dividing by $\sum_p m_p$ gives the bounds, and (i) gives the bound 1. \qed
\end{proof}

\begin{proof}[of Proposition~\ref{prop:complement}]
(i) By Lemma~\ref{lem:gini}, every $\Delta G_B(s)$ is unchanged, and so is any average of gains over labels. (ii) At each node the gain is unchanged, and if $p_L\neq p_R$ the child with the higher rate for $B$ becomes its sibling, so the gain is credited to the mirror predicate; if $p_L=p_R$ the gain is 0. The mirror map preserves feature and threshold or bin, so it is a bijection of aggregated keys with $g'_B(\bar v)=g_B(v)$. The activity rule depends only on the multiset of positive gains, so it commutes with this bijection. (iii) MDI is computed from these unchanged gains. Each tree's output for $B$ becomes one minus the original on the same structure, and SHAP values are linear in the model output and zero for constants, so they change sign only. Finally, $\phi(y_A,1-y_B)=-\phi(y_A,y_B)$. \qed
\end{proof}

\begin{proof}[of Proposition~\ref{prop:dissociation}]
(a) Take $K$ copies of a stump splitting $x_j$ at $t$, with two non-empty children and no bootstrap, and let $y_A=\mathbb{1}[x_j\leq t]$. Compare $y_B=y_A$ with $y_B=1-y_A$. By Proposition~\ref{prop:complement}, the summaries coincide, and $|\phi|=1$ in both. Each label has a single predicate with positive gain, which is therefore active. If $y_B=y_A$, both gains are credited to $(j,t,\leq)$, which lies on the only trace with positive mass for $B$, so SPF is 1. If $y_B=1-y_A$, the gain of $B$ is credited to $(j,t,>)$, so $C_{A,B}=\emptyset$ and SPF is 0 by Proposition~\ref{prop:range}.

(b) Let $n$ be a multiple of 4 and let the four values of $(y_A,y_B)$ occur equally often. With the single binary feature $z=\max(y_A,y_B)$, the only split has child rates 0 and $2/3$ for both labels, so both gains go to $(z,>)$, which is active for both labels and lies on the only trace with positive mass for $B$: SPF is 1. With the two binary features $z_A=y_A$ and $z_B=y_B$, a stump splits on $z_A$ or $z_B$, the two maximizers of the criterion. On $z_A$, the child rates of $B$ are equal, so by Lemma~\ref{lem:gini} $B$ receives no gain; symmetrically for $z_B$. Only predicates on $z_A$ can be active for $A$, and only predicates on $z_B$ for $B$, so $C_{A,B}=\emptyset$ and SPF is 0. \qed
\end{proof}

\begin{proof}[of Proposition~\ref{prop:lift}]
As in Proposition~\ref{prop:mass}(i), $W(v)$ and $F_\ell(v)$ are the in-bag mass and in-bag positive mass of $\ell$ on traces containing $v$, and by Eq.~\ref{eq:lift}, $\pi_\ell\lambda_\ell(v)=F_\ell(v)/W(v)$. (i) For every instance, $y_A+y_B=1+y_Ay_B-(1-y_A)(1-y_B)$; weighting by $c_{ki}$, summing over the in-bag instances on these traces and dividing by $W(v)$ gives the identity. (ii) Without co-occurrence, $F_{AB}(v)=0$ and $F_\emptyset(v)\geq0$. If $\lambda_A,\lambda_B>1$, then $\pi_A+\pi_B<\pi_A\lambda_A+\pi_B\lambda_B\leq1$. Moreover, $(\pi_A+\pi_B)\min\{\lambda_A,\lambda_B\}\leq\pi_A\lambda_A+\pi_B\lambda_B\leq1$. (iii) $W(v)\lambda_\ell(v)=F_\ell(v)/\pi_\ell$, and the traces through the elements of $V$ partition all traces, so $\sum_{v\in V}F_\ell(v)=\sum_r m(r,S_\ell)=\pi_\ell\sum_r w_r$. \qed
\end{proof}

\begin{proof}[of Proposition~\ref{prop:selection}]
At a node defined by fixing the values of some features, let $\sigma_j$ be a feature not yet fixed, with coefficient $c_j\in\{\alpha,\beta,0\}$ in the rate of a given label. By independence, the children receive half the mass each, and their rates differ by $2c_j$, the remaining terms averaging out equally in both. By Lemma~\ref{lem:gini}, the gain is $2c_j^2$, whatever the node. (i) Averaged over the two labels, the gain of $\sigma_s$ is $2\alpha^2$ and that of any private feature is $\beta^2$; $2\alpha^2>\beta^2$ because $\alpha>|\beta|/\sqrt2$. (ii) In the forest of $\ell$, private features of $\ell$ have gain $2\beta^2$, $\sigma_s$ has gain $2\alpha^2<2\beta^2$ and the features of the other label have gain 0. By induction on depth, a node at depth $d\leq m$ has fixed $d-1<m$ private features of $\ell$, so a private feature remains and is selected. The predicates of the two forests thus lie on disjoint features, and Proposition~\ref{prop:range}(ii) applies. (iii) The stump on $\sigma_s$ has child rates $\frac12\pm\alpha$ for both labels. Since $\alpha>0$, both gains are credited to $(s,>)$, the only predicate with positive gain, so $C_{A,B}=\{(s,>)\}$. The trace through it carries a share $\frac12(\frac12+\alpha)/\frac12=\frac12+\alpha$ of the mass of $B$. \qed
\end{proof}

\begin{proof}[of Proposition~\ref{prop:d1}]
Conditioning on features independent of $(x_0,x_1,x_2)$ leaves their joint distribution with $(y_0,y_1)$ unchanged, and the control labels have zero gain on $x_0,x_1,x_2$. Noise maps a rate $\rho$ to $\varepsilon+(1-2\varepsilon)\rho$, so by Lemma~\ref{lem:gini} every gain is multiplied by $(1-2\varepsilon)^2>0$, and we may take $\varepsilon=0$. For $y_0=\mathbb{1}[x_0>0,x_1>0]$ and a split of $x_0$ at $t>0$, let $a=P(x_0>t)<\frac12$. The child rates are $\frac12$ and $(\frac12-a)/(2(1-a))$, which differ by $1/(4(1-a))$, so the gain is $a/(8(1-a))<\frac18$. For $t<0$, let $b=P(x_0\leq t)<\frac12$. The child rates are 0 and $1/(4(1-b))$, so the gain is $b/(8(1-b))<\frac18$. At $t=0$ the gain is $\frac18$. The same holds for $y_1$, and for $y_0$ on $x_1$ and $y_1$ on $x_2$ by symmetry. Since $y_1$ is independent of $x_1$ and $y_0$ of $x_2$, the corresponding gains are 0. The summed gain over the two labels is thus uniquely maximized by $x_0$ at 0, with value $\frac14$, against at most $\frac18$ on $x_1$ or $x_2$. The BR statements are the ties at $\frac18$. \qed
\end{proof}

\section{Supplementary Analyses}
\label{app:analyses}

\paragraph{Degree-based variants of SPF.} Weighting each trace by the fraction of its predicates in $C_{A,B}$, instead of by the indicator of Eq.~\ref{eq:spf}, lowers the median SPF in synthetic scenario D1 (Section~\ref{sec:synth}) from .49 to .05, because most predicates on a trace are active for no label. Weighting it instead by the fraction of its $B$-active predicates that lie in $C_{A,B}$ avoids this dilution and is bounded above by SPF. This variant leaves the conclusions unchanged: the ordering of the synthetic scenarios is preserved (Kendall's $\tau=.97$ with SPF over repetitions), its rank correlation with SPF across the label pairs of \emph{emotions}, \emph{scene} and \emph{yeast} is at least .80 (median .98), and native forests at BR hyperparameters exceed BR in a similar share of pairs (64--92\%, against 63--90\% for SPF).

\paragraph{Global and local lift.} The local counterpart of the lift of predicate $v$ for label $\ell$ is $\sum_c w_cp_{c\ell}/\sum_c w_cp_{\mathrm{pa}(c)\ell}$ over the child nodes $c$ reached through $v$, with $\mathrm{pa}(c)$ the parent of $c$, and $w_c$ and $p_{c\ell}$ the in-bag mass and positive fraction of $c$. It exceeds 1 for 91\% of the predicate--label entries of active predicates in the native forests of \emph{emotions}, \emph{scene} and \emph{yeast}, since a predicate's gain is credited to the child in which the label's rate increases. The global and local readings classify shared predicates alike (co-enriching or separating) for 86\%, 96\% and 74\% of their mass on the three benchmarks; most disagreements are globally separating predicates that raise both rates locally but lead to a region enriched in one label and depleted in the other.

\paragraph{Containment.} For hierarchical labels, where a child label $C$ can only be positive together with a parent $P$, containment is the fraction of the child's decision mass carried by traces that contain at least one predicate supporting the parent. A predicate $v$ supports $P$ when at least 1\% of the parent's decision mass passes through it, $F_P(v)\geq0.01\sum_r m(r,S_P)$, where here $F_P(v)$ counts a trace once for each occurrence of $v$ on it. In scenario D4, where SPF is high in both directions, containment is .98 in the native forest and .82 under BR (medians over repetitions and over the two directions). Counting each trace once, as for the lifts, gives .980 and .818.

\paragraph{Sensitivity to the extraction settings.} Table~\ref{tab:robust} summarizes the 36 settings of Section~\ref{sec:res-robust} by benchmark: the minimum and maximum of the median SPF excess over the settings (Excess range); the number of settings in which its sign equals that at the reference setting (Same sign); the median Spearman correlation between the pair rankings under the 27 quantile-bin settings and under the reference setting ($\rho$); and, at the reference setting, the median Jaccard excess (Jacc.), the mean number of active predicates per label ($|\mathcal{A}_\ell|$, native/BR) and the fraction of ordered pairs whose shared active set is empty (Empty, native/BR).

\begin{table}[!htbp]
\caption{Sensitivity of the native--BR comparison to the 36 extraction settings; columns as defined in the text.}
\centering
\scriptsize
\setlength{\tabcolsep}{3pt}
\begin{tabular}{lrrrrrr}
\toprule
Dataset & Excess range & Same sign & $\rho$ & Jacc. & $|\mathcal{A}_\ell|$ & Empty\\
\midrule
emotions & .044 to .620 & 36 & .92 & .032 & 340/96 & .13/.53\\
scene    & .004 to .531 & 36 & .93 & .006 & 360/283 & .20/.53\\
yeast    & .089 to .630 & 36 & .78 & .033 & 878/500 & .00/.07\\
image    & .059 to .569 & 36 & .85 & .038 & 872/418 & .00/.05\\
birds    & .159 to .641 & 36 & .81 & .049 & 323/125 & .00/.04\\
genbase  & .000 to .377 & 24 & .01 & .106 & 4.4/3.9 & .23/.86\\
enron    & .034 to .327 & 36 & .57 & .059 & 174/115 & .00/.13\\
slashdot & .147 to .747 & 36 & .58 & .013 & 112/28 & .21/.71\\
langlog  & $-.033$ to .095 & 24 & .48 & $-.016$ & 254/249 & .00/.00\\
\bottomrule
\end{tabular}
\label{tab:robust}
\end{table}

\paragraph{Coverage and stability of the contracted views.} Table~\ref{tab:views} details the analysis of Section~\ref{sec:res-flags} for \texttt{sad.lonely}$\to$\texttt{quiet.still} (\emph{emotions}) and \texttt{white}$\to$\texttt{blue} (\emph{flags}), each view displaying 14 predicates. Coverage is the fraction of each label's decision mass (source/target) on traces through a displayed predicate, and Omitted the fraction of a class's leaf mass on traces through none. Jaccard is the index between the displayed sets of two repetitions, with predicates matched by feature, direction and overlapping threshold bins, or by feature and direction only. Recurring counts the predicates of the figure that are displayed, or shared and active, in at least five of the other nine repetitions, and Lift agreement is the mass-weighted fraction of comparisons in which a shared active predicate has the same lift classification in another repetition.

\begin{table}[!htbp]
\caption{Coverage and stability of the contracted views of Figs.~\ref{fig:msdpg-graph} and~\ref{fig:msdpg-flags}: medians over repetitions, first repetition in brackets; rows as defined in the text.}
\centering
\scriptsize
\setlength{\tabcolsep}{4pt}
\begin{tabular}{lrr}
\toprule
 & \emph{emotions} & \emph{flags}\\
\midrule
SPF & .88 [.92] & .97 [.97]\\
Shared active & 75 [131] & 30.5 [31]\\
Co-enriching & 74.5 [130] & 17.5 [17]\\
Coverage & .44/.52 [.37/.44] & .79/.82 [.77/.82]\\
Omitted, both & .42 [.51] & .16 [.16]\\
Omitted, source only & .81 [.84] & .30 [.34]\\
Omitted, target only & .61 [.71] & .30 [.36]\\
Omitted, neither & .88 [.89] & .43 [.42]\\
Jaccard (range) & .27 (.08--.47) & .33 (.17--.56)\\
\quad feature, direction & .46 (.25--.70) & .35 (.17--.56)\\
Recurring, displayed & 4 of 14 & 6 of 14\\
Recurring, shared active & 9 of 14 & 8 of 14\\
Lift agreement & .997 & .65\\
\bottomrule
\end{tabular}
\label{tab:views}
\end{table}

\paragraph{Model size.} Table~\ref{tab:size} gives the size of the native and BR models, as medians over repetitions.

\begin{table}[!htbp]
\caption{Size of the native and BR models: number of trees, total number of nodes, and mean and maximum tree depth (medians over repetitions).}
\centering
\scriptsize
\setlength{\tabcolsep}{4.5pt}
\begin{tabular}{lrrrrrrrr}
\toprule
 & \multicolumn{4}{c}{Native} & \multicolumn{4}{c}{BR}\\
\cmidrule(lr){2-5}\cmidrule(lr){6-9}
Dataset & Trees & Nodes & Mean d. & Max d. & Trees & Nodes & Mean d. & Max d.\\
\midrule
emotions & 200 & 49,521 & 13.4 & 20 & 1,200 & 56,801 & 7.7 & 8\\
scene    & 200 & 101,766 & 18.6 & 28 & 1,200 & 110,575 & 8.0 & 8\\
yeast    & 200 & 225,598 & 21.7 & 32 & 1,600 & 199,421 & 8.0 & 8\\
image    & 200 & 133,961 & 19.3 & 27 & 1,000 & 117,063 & 8.0 & 8\\
birds    & 200 & 39,212 & 15.9 & 28 & 1,600 & 51,603 & 7.6 & 12\\
genbase  & 200 & 11,235 & 19.7 & 29 & 1,600 & 38,929 & 9.6 & 30\\
enron    & 200 & 215,393 & 59.3 & 77 & 1,600 & 317,434 & 16.0 & 16\\
slashdot & 200 & 431,688 & 236.1 & 282 & 1,600 & 80,615 & 12.0 & 12\\
langlog  & 200 & 115,456 & 31.8 & 48 & 1,600 & 186,827 & 14.5 & 28\\
\bottomrule
\end{tabular}
\label{tab:size}
\end{table}

\section{Atlas of Explanations}
\label{app:atlas}

This appendix applies the two levels of the explanation to the native forest of the first repetition of every benchmark, with the reference extraction setting. In each figure, the left panel is the SPF matrix (rows: source label $A$; columns: target label $B$), with the ordered pair of highest SPF outlined. The right panel places every shared active predicate of that pair by its two lifts, on logarithmic axes: predicates in the upper right quadrant are co-enriching (green), those in the upper left and lower right quadrants separating (magenta), and those in the lower left co-depleting (grey); point size is proportional to the decision mass of the two labels on the traces through the predicate. The six predicates with the largest mass are numbered and listed below the panels with their lifts, their share of the shared decision mass, and the number of the other nine repetitions in which a matching predicate (same feature and direction, overlapping threshold bin) is shared and active for the same pair (Rec.). Predicates are written with the mean threshold of the splits they merge, and features with two values as present or absent. The SPF of the selected pair in the other repetitions and the share of separating mass are given in the text.

\begin{figure}[!htbp]
\centering
\includegraphics[width=\textwidth]{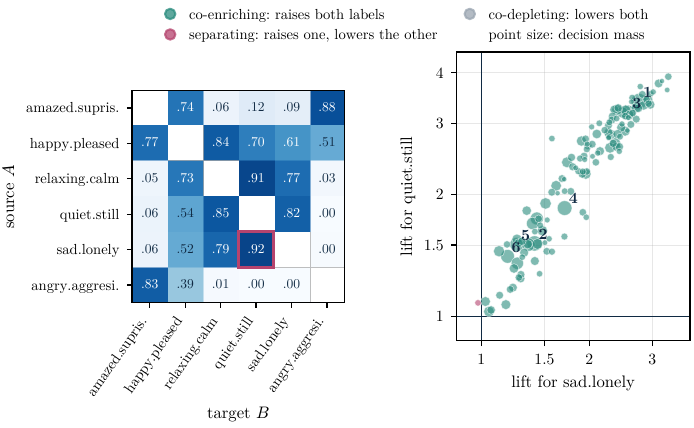}

\vspace{2pt}
\scriptsize
\begin{tabular}{@{}rlrrrr@{}}
\toprule
 & Shared active predicate & $\lambda_A$ & $\lambda_B$ & Mass (\%) & Rec.\\
\midrule
1 & Mean\_Acc1298\_Mean\_Mem40\_Rolloff $\leq$ 0.104 & 2.75 & 3.40 & 2.9 & 9\\
2 & Std\_Acc1298\_Mean\_Mem40\_MFCC\_11 $>$ 0.234 & 1.41 & 1.51 & 2.7 & 7\\
3 & Mean\_Acc1298\_Mean\_Mem40\_Centroid $\leq$ 0.0532 & 2.57 & 3.19 & 2.5 & 8\\
4 & Std\_Acc1298\_Mean\_Mem40\_MFCC\_8 $>$ 0.26 & 1.71 & 1.85 & 2.5 & 1\\
5 & Std\_Acc1298\_Mean\_Mem40\_MFCC\_0 $>$ 2.7 & 1.26 & 1.51 & 2.3 & 1\\
6 & Std\_Acc1298\_Std\_Mem40\_MFCC\_11 $>$ 0.0964 & 1.18 & 1.41 & 2.1 & 3\\
\bottomrule
\end{tabular}
\caption{\emph{emotions}: SPF matrix and lifts of the shared active predicates of \texttt{sad.lonely}$\to$\texttt{quiet.still}.}
\label{fig:atlas-emotions}
\end{figure}

\paragraph{\emph{emotions}.} Selected pair \texttt{sad.lonely}$\to$\texttt{quiet.still}: SPF .92 (.67 to .93 in the other repetitions), 131 shared active predicates, less than 1\% of their decision mass on separating predicates (Fig.~\ref{fig:atlas-emotions}). The highest SPF links the two quiet emotions, and almost all of the shared mass lies on co-enriching predicates: low mean spectral rolloff and centroid raise both labels about three times above their forest-wide rates and recur in eight or nine of the other repetitions, whereas the variability conditions on individual MFCCs recur less often.

\begin{figure}[!htbp]
\centering
\includegraphics[width=\textwidth]{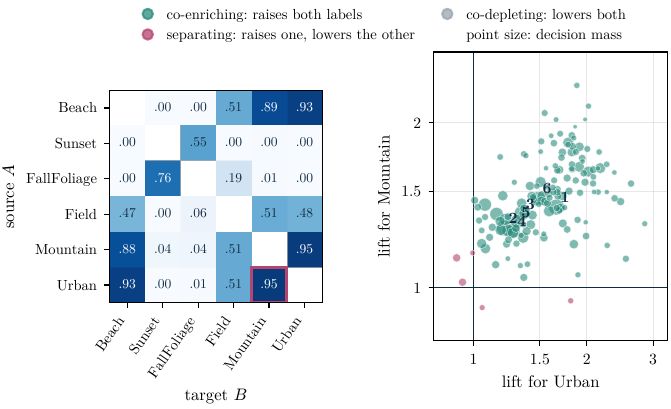}

\vspace{2pt}
\scriptsize
\begin{tabular}{@{}rlrrrr@{}}
\toprule
 & Shared active predicate & $\lambda_A$ & $\lambda_B$ & Mass (\%) & Rec.\\
\midrule
1 & Att233 $\leq$ 0.74 & 1.66 & 1.40 & 2.9 & 8\\
2 & Att102 $\leq$ 0.364 & 1.21 & 1.29 & 2.8 & 9\\
3 & Att103 $\leq$ 0.358 & 1.34 & 1.37 & 2.6 & 9\\
4 & Att115 $\leq$ 0.438 & 1.27 & 1.27 & 2.2 & 9\\
5 & Att107 $\leq$ 0.37 & 1.31 & 1.32 & 2.1 & 9\\
6 & Att207 $\leq$ 0.566 & 1.49 & 1.46 & 2.1 & 9\\
\bottomrule
\end{tabular}
\caption{\emph{scene}: SPF matrix and lifts of the shared active predicates of \texttt{Urban}$\to$\texttt{Mountain}.}
\label{fig:atlas-scene}
\end{figure}

\paragraph{\emph{scene}.} Selected pair \texttt{Urban}$\to$\texttt{Mountain}: SPF .95 (.95 to .97 in the other repetitions), 144 shared active predicates, 2\% of their decision mass on separating predicates (Fig.~\ref{fig:atlas-scene}). The features are anonymous colour and texture attributes, so the predicates can be compared across labels and repetitions but not read semantically. Their lifts lie between 1.2 and 1.7 for both labels, few shared predicates are separating, and the six largest recur in eight or nine of the other repetitions.

\begin{figure}[!htbp]
\centering
\includegraphics[width=\textwidth]{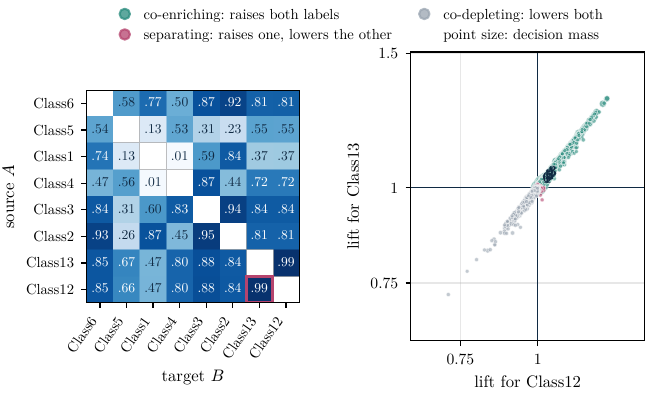}

\vspace{2pt}
\scriptsize
\begin{tabular}{@{}rlrrrr@{}}
\toprule
 & Shared active predicate & $\lambda_A$ & $\lambda_B$ & Mass (\%) & Rec.\\
\midrule
1 & Att88 $\leq$ 0.0919 & 1.00 & 1.00 & 1.1 & 9\\
2 & Att88 $\leq$ 0.123 & 1.01 & 1.01 & 0.9 & 9\\
3 & Att3 $>$ -0.105 & 1.02 & 1.02 & 0.7 & 7\\
4 & Att96 $\leq$ 0.0766 & 1.02 & 1.02 & 0.7 & 9\\
5 & Att98 $\leq$ 0.136 & 1.01 & 1.01 & 0.6 & 8\\
6 & Att96 $\leq$ -0.0214 & 1.00 & 1.00 & 0.5 & 7\\
\bottomrule
\end{tabular}
\caption{\emph{yeast}: SPF matrix and lifts of the shared active predicates of \texttt{Class12}$\to$\texttt{Class13}.}
\label{fig:atlas-yeast}
\end{figure}

\paragraph{\emph{yeast}.} Selected pair \texttt{Class12}$\to$\texttt{Class13}: SPF .99 (.99 to .99 in the other repetitions), 1,545 shared active predicates, 3\% of their decision mass on separating predicates (Fig.~\ref{fig:atlas-yeast}). Both labels of the pair are positive for three quarters of the instances. SPF is .99 and 1,545 predicates are shared, but the largest have lifts between 1.00 and 1.02 for both labels: the two labels lie on almost the same traces without being enriched by any of them. A high SPF combined with lifts close to 1 thus indicates sharing that the lifts cannot differentiate.

\begin{figure}[!htbp]
\centering
\includegraphics[width=\textwidth]{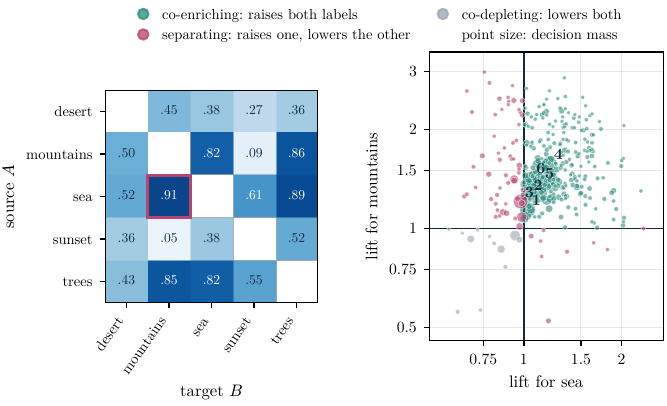}

\vspace{2pt}
\scriptsize
\begin{tabular}{@{}rlrrrr@{}}
\toprule
 & Shared active predicate & $\lambda_A$ & $\lambda_B$ & Mass (\%) & Rec.\\
\midrule
1 & Feature4 $\leq$ 170 & 1.03 & 1.14 & 3.5 & 4\\
2 & Feature6 $\leq$ 156 & 1.04 & 1.27 & 3.2 & 4\\
3 & Feature1 $\leq$ 215 & 0.98 & 1.21 & 2.9 & 7\\
4 & Feature9 $\leq$ 105 & 1.20 & 1.58 & 2.8 & 9\\
5 & Feature19 $>$ 91.8 & 1.13 & 1.37 & 2.7 & 9\\
6 & Feature21 $>$ 89.3 & 1.06 & 1.42 & 2.6 & 9\\
\bottomrule
\end{tabular}
\caption{\emph{image}: SPF matrix and lifts of the shared active predicates of \texttt{sea}$\to$\texttt{mountains}.}
\label{fig:atlas-image}
\end{figure}

\paragraph{\emph{image}.} Selected pair \texttt{sea}$\to$\texttt{mountains}: SPF .91 (.71 to .90 in the other repetitions), 335 shared active predicates, 15\% of their decision mass on separating predicates (Fig.~\ref{fig:atlas-image}). Most of the shared mass lies on predicates on colour features that raise \texttt{mountains} more than \texttt{sea}; 15\% lies on separating predicates. The largest predicates recur in four to nine of the other repetitions.

\begin{figure}[!htbp]
\centering
\includegraphics[width=\textwidth]{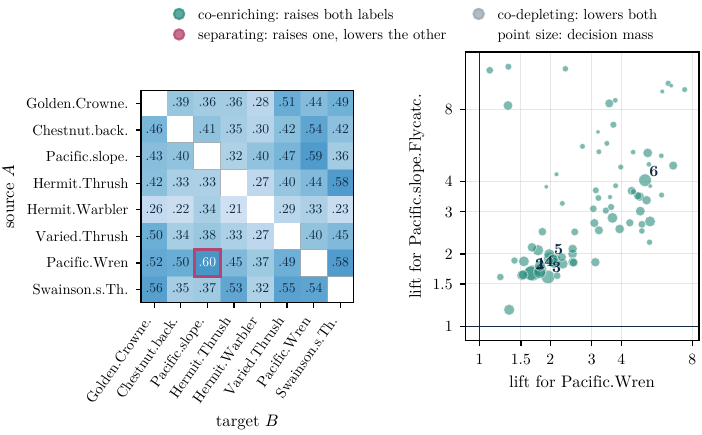}

\vspace{2pt}
\scriptsize
\begin{tabular}{@{}rlrrrr@{}}
\toprule
 & Shared active predicate & $\lambda_A$ & $\lambda_B$ & Mass (\%) & Rec.\\
\midrule
1 & hasSegments present & 1.68 & 1.66 & 4.9 & 9\\
2 & mean\_rect\_height $>$ 5.11 & 1.66 & 1.66 & 4.0 & 9\\
3 & mean\_rect\_width $>$ 19.1 & 1.95 & 1.60 & 3.4 & 7\\
4 & audio.ssd68 $>$ 0.74 & 1.81 & 1.71 & 3.0 & 9\\
5 & segments $>$ 1.5 & 1.99 & 1.92 & 3.0 & 9\\
6 & audio.ssd44 $>$ 0.00321 & 5.05 & 4.05 & 3.0 & 6\\
\bottomrule
\end{tabular}
\caption{\emph{birds}: SPF matrix and lifts of the shared active predicates of \texttt{Pacific.Wren}$\to$\texttt{Pacific.slope.Flycatcher}.}
\label{fig:atlas-birds}
\end{figure}

\paragraph{\emph{birds}.} Selected pair \texttt{Pacific.Wren}$\to$\texttt{Pacific.slope.Flycatcher}: SPF .60 (.51 to .77 in the other repetitions), 79 shared active predicates, none of their decision mass on separating predicates (Fig.~\ref{fig:atlas-birds}). The largest shared predicates are conditions on the segments detected in a recording, such as their presence, number and size, together with two audio descriptors; none is separating, so the forest raises the probability of both species where a segment is detected. Four of the six recur in all nine other repetitions.

\begin{figure}[!htbp]
\centering
\includegraphics[width=\textwidth]{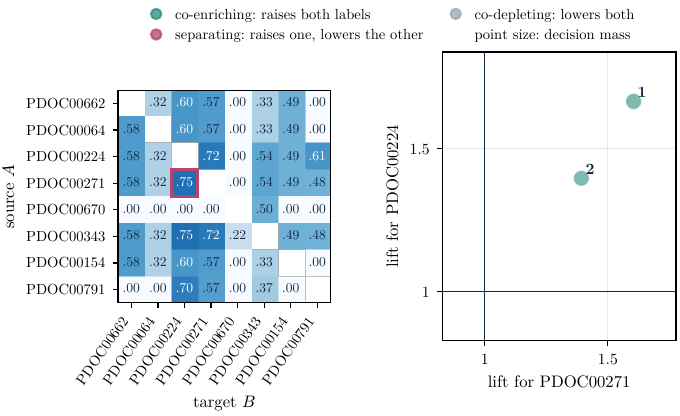}

\vspace{2pt}
\scriptsize
\begin{tabular}{@{}rlrrrr@{}}
\toprule
 & Shared active predicate & $\lambda_A$ & $\lambda_B$ & Mass (\%) & Rec.\\
\midrule
1 & PS01031 absent & 1.63 & 1.72 & 50.8 & 9\\
2 & PS50072 absent & 1.37 & 1.38 & 49.2 & 1\\
\bottomrule
\end{tabular}
\caption{\emph{genbase}: SPF matrix and lifts of the shared active predicates of \texttt{PDOC00271}$\to$\texttt{PDOC00224}.}
\label{fig:atlas-genbase}
\end{figure}

\paragraph{\emph{genbase}.} Selected pair \texttt{PDOC00271}$\to$\texttt{PDOC00224}: SPF .75 (.56 to .69 in the other repetitions), 2 shared active predicates, none of their decision mass on separating predicates (Fig.~\ref{fig:atlas-genbase}). Only two predicates are shared, both the absence of a protein motif, and both raise the two families: the forest decides these labels partly through the absence of these motifs. The first recurs in all other repetitions, the second in one.

\begin{figure}[!htbp]
\centering
\includegraphics[width=\textwidth]{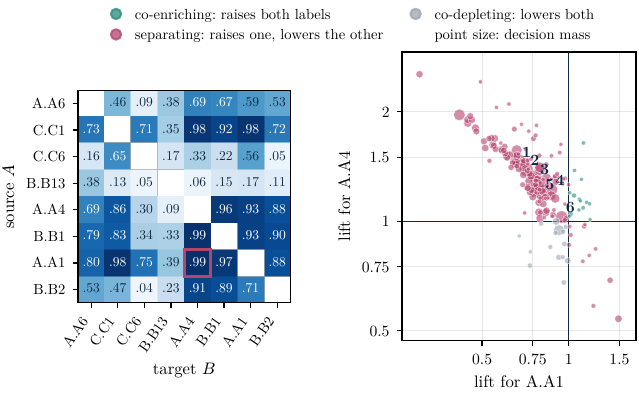}

\vspace{2pt}
\scriptsize
\begin{tabular}{@{}rlrrrr@{}}
\toprule
 & Shared active predicate & $\lambda_A$ & $\lambda_B$ & Mass (\%) & Rec.\\
\midrule
1 & attached absent & 0.67 & 1.46 & 3.6 & 7\\
2 & draft absent & 0.71 & 1.39 & 2.7 & 9\\
3 & doc absent & 0.77 & 1.31 & 2.2 & 9\\
4 & pmto absent & 0.87 & 1.22 & 2.2 & 9\\
5 & original absent & 0.80 & 1.19 & 2.1 & 0\\
6 & enron present & 0.94 & 1.03 & 2.1 & 5\\
\bottomrule
\end{tabular}
\caption{\emph{enron}: SPF matrix and lifts of the shared active predicates of \texttt{A.A1}$\to$\texttt{A.A4}.}
\label{fig:atlas-enron}
\end{figure}

\paragraph{\emph{enron}.} Selected pair \texttt{A.A1}$\to$\texttt{A.A4}: SPF .99 (.96 to .99 in the other repetitions), 176 shared active predicates, 94\% of their decision mass on separating predicates (Fig.~\ref{fig:atlas-enron}). Although SPF is .99, 94\% of the shared mass lies on separating predicates: the absence of words such as \emph{attached}, \emph{draft} and \emph{doc} raises \texttt{A.A4} and lowers \texttt{A.A1}. The forest decides the two labels through the same conditions, but uses them to choose one rather than the other, which SPF alone does not reveal.

\begin{figure}[!htbp]
\centering
\includegraphics[width=\textwidth]{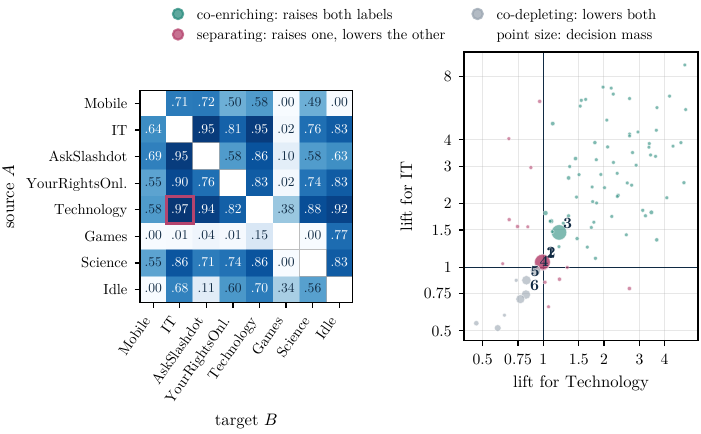}

\vspace{2pt}
\scriptsize
\begin{tabular}{@{}rlrrrr@{}}
\toprule
 & Shared active predicate & $\lambda_A$ & $\lambda_B$ & Mass (\%) & Rec.\\
\midrule
1 & game absent & 0.99 & 1.05 & 19.5 & 9\\
2 & games absent & 0.99 & 1.06 & 18.9 & 4\\
3 & writes present & 1.20 & 1.46 & 18.1 & 9\\
4 & linux absent & 0.91 & 0.96 & 7.9 & 9\\
5 & software absent & 0.82 & 0.87 & 6.0 & 6\\
6 & anonymous absent & 0.82 & 0.74 & 5.9 & 9\\
\bottomrule
\end{tabular}
\caption{\emph{slashdot}: SPF matrix and lifts of the shared active predicates of \texttt{Technology}$\to$\texttt{IT}.}
\label{fig:atlas-slashdot}
\end{figure}

\paragraph{\emph{slashdot}.} Selected pair \texttt{Technology}$\to$\texttt{IT}: SPF .97 (.96 to .98 in the other repetitions), 88 shared active predicates, 41\% of their decision mass on separating predicates (Fig.~\ref{fig:atlas-slashdot}). The two largest shared predicates, the absence of \emph{game} and \emph{games}, have lifts close to 1 and carry little information about either label, whereas the presence of \emph{writes} raises both; 41\% of the shared mass lies on separating predicates.

\begin{figure}[!htbp]
\centering
\includegraphics[width=\textwidth]{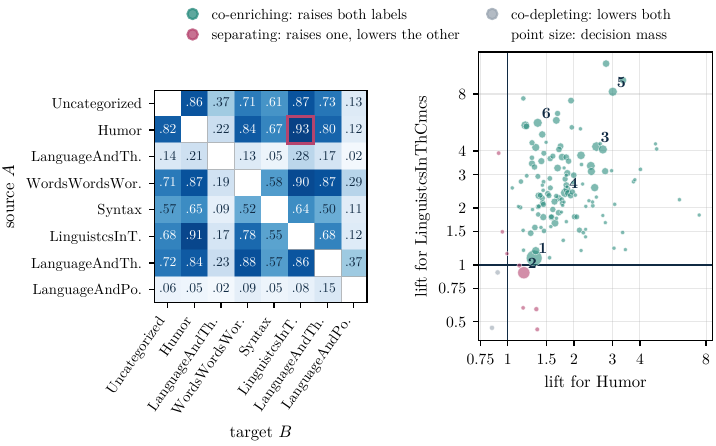}

\vspace{2pt}
\scriptsize
\begin{tabular}{@{}rlrrrr@{}}
\toprule
 & Shared active predicate & $\lambda_A$ & $\lambda_B$ & Mass (\%) & Rec.\\
\midrule
1 & tok242 absent & 1.32 & 1.09 & 7.5 & 4\\
2 & tok642 absent & 1.18 & 0.91 & 4.2 & 0\\
3 & tok438 absent & 2.53 & 4.22 & 2.5 & 9\\
4 & tok906 absent & 1.82 & 2.40 & 2.3 & 9\\
5 & tok413 absent & 3.01 & 8.22 & 2.2 & 9\\
6 & tok154 absent & 1.37 & 5.64 & 2.1 & 9\\
\bottomrule
\end{tabular}
\caption{\emph{langlog}: SPF matrix and lifts of the shared active predicates of \texttt{Humor}$\to$\texttt{LinguistcsInThCmcs}.}
\label{fig:atlas-langlog}
\end{figure}

\paragraph{\emph{langlog}.} Selected pair \texttt{Humor}$\to$\texttt{LinguistcsInThCmcs}: SPF .93 (.71 to .92 in the other repetitions), 138 shared active predicates, 6\% of their decision mass on separating predicates (Fig.~\ref{fig:atlas-langlog}). The features are anonymized tokens, so the predicates cannot be read. Most shared predicates are co-enriching, and several of them raise both labels strongly and recur in all other repetitions; the example shows the limit that uninterpretable features set on a predicate-level explanation.

\bibliographystyle{splncs04}
\bibliography{references}

\begin{thebibliography}{10}
\providecommand{\url}[1]{\texttt{#1}}
\providecommand{\urlprefix}{URL }
\providecommand{\doi}[1]{https://doi.org/#1}

\bibitem{adadi2018peeking}
Adadi, A., Berrada, M.: Peeking inside the black-box: A survey on explainable
  artificial intelligence ({XAI}). IEEE Access  \textbf{6},  52138--52160
  (2018)

\bibitem{arrighi2024dpg}
Arrighi, L., Pennella, L., Marques~Tavares, G., Barbon~Junior, S.: Decision
  predicate graphs: Enhancing interpretability in tree ensembles. In: World
  Conference on Explainable Artificial Intelligence. pp. 311--332. Springer
  Nature Switzerland (2024). \doi{10.1007/978-3-031-63797-1_16}

\bibitem{bardos2023mtr}
Bardos, A., Mylonas, N., Mollas, I., Tsoumakas, G.: Local interpretability of
  random forests for multi-target regression. arXiv preprint arXiv:2303.16506
  (2023)

\bibitem{blockeel1998pct}
Blockeel, H., De~Raedt, L., Ramon, J.: Top-down induction of clustering trees.
  In: Proceedings of the 15th International Conference on Machine Learning
  (ICML). pp. 55--63 (1998)

\bibitem{breiman2001random}
Breiman, L.: Random forests. Machine Learning  \textbf{45}(1),  5--32 (2001).
  \doi{10.1023/A:1010933404324}

\bibitem{briggs2013birds}
Briggs, F., et~al.: The 9th annual {MLSP} competition: New methods for acoustic
  classification of multiple simultaneous bird species in a noisy environment.
  In: IEEE International Workshop on Machine Learning for Signal Processing
  (MLSP). pp.~1--8 (2013)

\bibitem{ceschin2026isolation}
Ceschin, M., Arrighi, L., Longo, L., Barbon~Junior, S.: Extending decision
  predicate graphs for comprehensive explanation of isolation forest. In:
  Explainable Artificial Intelligence. Communications in Computer and
  Information Science, vol.~2577, pp. 271--293. Springer (2026).
  \doi{10.1007/978-3-032-08324-1_12}

\bibitem{chen2020true}
Chen, H., Janizek, J.D., Lundberg, S., Lee, S.I.: True to the model or true to
  the data? arXiv preprint arXiv:2006.16234  (2020)

\bibitem{chen2021interpretation}
Chen, S.: Interpretation of multi-label classification models using shapley
  values. CoRR  \textbf{abs/2104.10505} (2021)

\bibitem{dembczynski2012dependence}
Dembczy\'nski, K., Waegeman, W., Cheng, W., H\"ullermeier, E.: On label
  dependence and loss minimization in multi-label classification. Machine
  Learning  \textbf{88}(1--2),  5--45 (2012). \doi{10.1007/s10994-012-5285-8}

\bibitem{diplaris2005protein}
Diplaris, S., Tsoumakas, G., Mitkas, P.A., Vlahavas, I.: Protein classification
  with multiple algorithms. In: Advances in Informatics, 10th Panhellenic
  Conference on Informatics (PCI 2005). LNCS, vol.~3746, pp. 448--456. Springer
  (2005)

\bibitem{feng2026semgnn}
Feng, Y., Tang, Y., Shi, M., Zhu, X.: Self-explainable multi-label graph neural
  network for correlated evidence attribution. arXiv preprint arXiv:2608.27574
  (2026). \doi{10.48550/arXiv.2608.27574}

\bibitem{friedman2008rulefit}
Friedman, J.H., Popescu, B.E.: Predictive learning via rule ensembles. The
  Annals of Applied Statistics  \textbf{2}(3),  916--954 (2008).
  \doi{10.1214/07-AOAS148}

\bibitem{huang2015llsf}
Huang, J., Li, G., Huang, Q., Wu, X.: Learning label specific features for
  multi-label classification. In: Proceedings of the IEEE International
  Conference on Data Mining (ICDM). pp. 181--190 (2015)

\bibitem{klimt2004enron}
Klimt, B., Yang, Y.: The {Enron} corpus: A new dataset for email classification
  research. In: Machine Learning: ECML 2004. LNCS, vol.~3201, pp. 217--226.
  Springer (2004)

\bibitem{kocev2013ensembles}
Kocev, D., Vens, C., Struyf, J., D{\v{z}}eroski, S.: Tree ensembles for
  predicting structured outputs. Pattern Recognition  \textbf{46}(3),  817--833
  (2013)

\bibitem{lundberg2020trees}
Lundberg, S.M., Erion, G., Chen, H., et~al.: From local explanations to global
  understanding with explainable {AI} for trees. Nature Machine Intelligence
  \textbf{2},  56--67 (2020). \doi{10.1038/s42256-019-0138-9}

\bibitem{lundberg2017shap}
Lundberg, S.M., Lee, S.I.: A unified approach to interpreting model
  predictions. In: Advances in Neural Information Processing Systems. vol.~30
  (2017)

\bibitem{mastelini2019motc}
Mastelini, S.M., da~Costa, V.G.T., Santana, E.J., Nakano, F.K., Guido, R.C.,
  Cerri, R., Barbon~Junior, S.: Multi-output tree chaining: An interpretative
  modelling and lightweight multi-target approach. Journal of Signal Processing
  Systems  \textbf{91}(2),  191--215 (2019). \doi{10.1007/s11265-018-1376-5}

\bibitem{mylonas2022lionforestsml}
Mylonas, N., Mollas, I., Bassiliades, N., Tsoumakas, G.: Local multi-label
  explanations for random forest. In: Machine Learning and Principles and
  Practice of Knowledge Discovery in Databases. ECML PKDD 2022. Communications
  in Computer and Information Science, vol.~1752, pp. 369--384. Springer
  (2022). \doi{10.1007/978-3-031-23618-1_25}

\bibitem{nauta2023anecdotal}
Nauta, M., Trienes, J., Pathak, S., Nguyen, E., Peters, M., Schmitt, Y.,
  Schl{\"o}tterer, J., van Keulen, M., Seifert, C.: From anecdotal evidence to
  quantitative evaluation methods: A systematic review on evaluating
  explainable {AI}. ACM Computing Surveys  \textbf{55}(13s),  295:1--295:42
  (2023)

\bibitem{panigutti2020marlena}
Panigutti, C., Guidotti, R., Monreale, A., Pedreschi, D.: Explaining
  multi-label black-box classifiers for health applications. In: Explainable
  and Interpretable Models in Computer Vision and Machine Learning, Studies in
  Computational Intelligence, vol.~843, pp. 97--110. Springer (2020).
  \doi{10.1007/978-3-030-24409-5_9}

\bibitem{pedregosa2011sklearn}
Pedregosa, F., Varoquaux, G., Gramfort, A., Michel, V., Thirion, B., Grisel,
  O., Blondel, M., Prettenhofer, P., Weiss, R., Dubourg, V., Vanderplas, J.,
  Passos, A., Cournapeau, D., Brucher, M., Perrot, M., Duchesnay, {\'E}.:
  Scikit-learn: Machine learning in {P}ython. Journal of Machine Learning
  Research  \textbf{12},  2825--2830 (2011)

\bibitem{petkovic2020feature}
Petkovi\'c, M., Kocev, D., D\v{z}eroski, S.: Feature ranking for multi-target
  regression. Machine Learning  \textbf{109},  1179--1204 (2020).
  \doi{10.1007/s10994-019-05829-8}

\bibitem{read2011chains}
Read, J., Pfahringer, B., Holmes, G., Frank, E.: Classifier chains for
  multi-label classification. Machine Learning  \textbf{85}(3),  333--359
  (2011). \doi{10.1007/s10994-011-5256-5}

\bibitem{tabia2019multilabel}
Tabia, K.: Towards explainable multi-label classification. In: 2019 IEEE 31st
  International Conference on Tools with Artificial Intelligence (ICTAI). pp.
  1088--1095 (2019). \doi{10.1109/ICTAI.2019.00152}

\bibitem{trohidis2008emotions}
Trohidis, K., Tsoumakas, G., Kalliris, G., Vlahavas, I.: Multi-label
  classification of music into emotions. In: Proceedings of the 9th
  International Conference on Music Information Retrieval (ISMIR). pp. 325--330
  (2008)

\bibitem{tsoumakas2007overview}
Tsoumakas, G., Katakis, I.: Multi-label classification: An overview.
  International Journal of Data Warehousing and Mining  \textbf{3}(3),  1--13
  (2007). \doi{10.4018/jdwm.2007070101}

\bibitem{tzanetakis2002genre}
Tzanetakis, G., Cook, P.: Musical genre classification of audio signals. IEEE
  Transactions on Speech and Audio Processing  \textbf{10}(5),  293--302 (2002)

\bibitem{vanschoren2014openml}
Vanschoren, J., van Rijn, J.N., Bischl, B., Torgo, L.: {OpenML}: Networked
  science in machine learning. SIGKDD Explorations  \textbf{15}(2),  49--60
  (2014)

\bibitem{zhang2015lift}
Zhang, M.L., Wu, L.: {LIFT}: Multi-label learning with label-specific features.
  IEEE Transactions on Pattern Analysis and Machine Intelligence
  \textbf{37}(1),  107--120 (2015)

\bibitem{zhang2007mlknn}
Zhang, M.L., Zhou, Z.H.: {ML-KNN}: A lazy learning approach to multi-label
  learning. Pattern Recognition  \textbf{40}(7),  2038--2048 (2007)

\bibitem{zhang2014multilabel}
Zhang, M.L., Zhou, Z.H.: A review on multi-label learning algorithms. IEEE
  Transactions on Knowledge and Data Engineering  \textbf{26}(8),  1819--1837
  (2014). \doi{10.1109/TKDE.2013.39}

\end{thebibliography}
\end{document}